\documentclass[10pt,twocolumn,letterpaper]{article}

\usepackage{cvpr}              % To produce the CAMERA-READY version
\usepackage{booktabs}
\usepackage{multirow}
\usepackage{float}
\usepackage{graphicx}   % for \rotatebox
\usepackage[dvipsnames]{xcolor}
\usepackage{amssymb}    % for \checkmark
\usepackage{utfsym}
\usepackage{chemformula}
\usepackage{svg}      % 提供 \includesvg
\usepackage{placeins}   % 提供 \FloatBarrier

\definecolor{cvprblue}{rgb}{0.21,0.49,0.74}
\usepackage[pagebackref,breaklinks,colorlinks,allcolors=cvprblue]{hyperref}

\def\paperID{3006} % *** Enter the Paper ID here
\def\confName{CVPR}
\def\confYear{2026}

\title{RoboTidy : A 3D Gaussian Splatting Household Tidying Benchmark for Embodied Navigation and Action}

\author{
Xiaoquan Sun$^{1}$\footnotemark[1]\quad 
Ruijian Zhang$^{1}$\footnotemark[1]\quad 
Kang Pang$^{1}$ \quad
Bingchen Miao$^{4}$ \quad
Yuxiang Tan$^{1}$ \\
Zhen Yang$^{2}$ \quad
Ming Li$^{5}$ \quad
Jiayu Chen$^{2,3}$\footnotemark[2] \\[0.3em]
$^{1}$Huazhong University of Science and Technology \quad
$^{2}$The University of Hong Kong \\
$^{3}$INFIFORCE Intelligent Technology Co., Ltd. \quad
$^{4}$Zhejiang University \quad
$^{5}$Guangming Lab, Shenzhen
}

\begin{document}

\twocolumn[{%
\renewcommand\twocolumn[1][]{#1}%
\maketitle
\begin{center}
    \centering
    \captionsetup{type=figure}
    \includegraphics[width=0.99\linewidth]{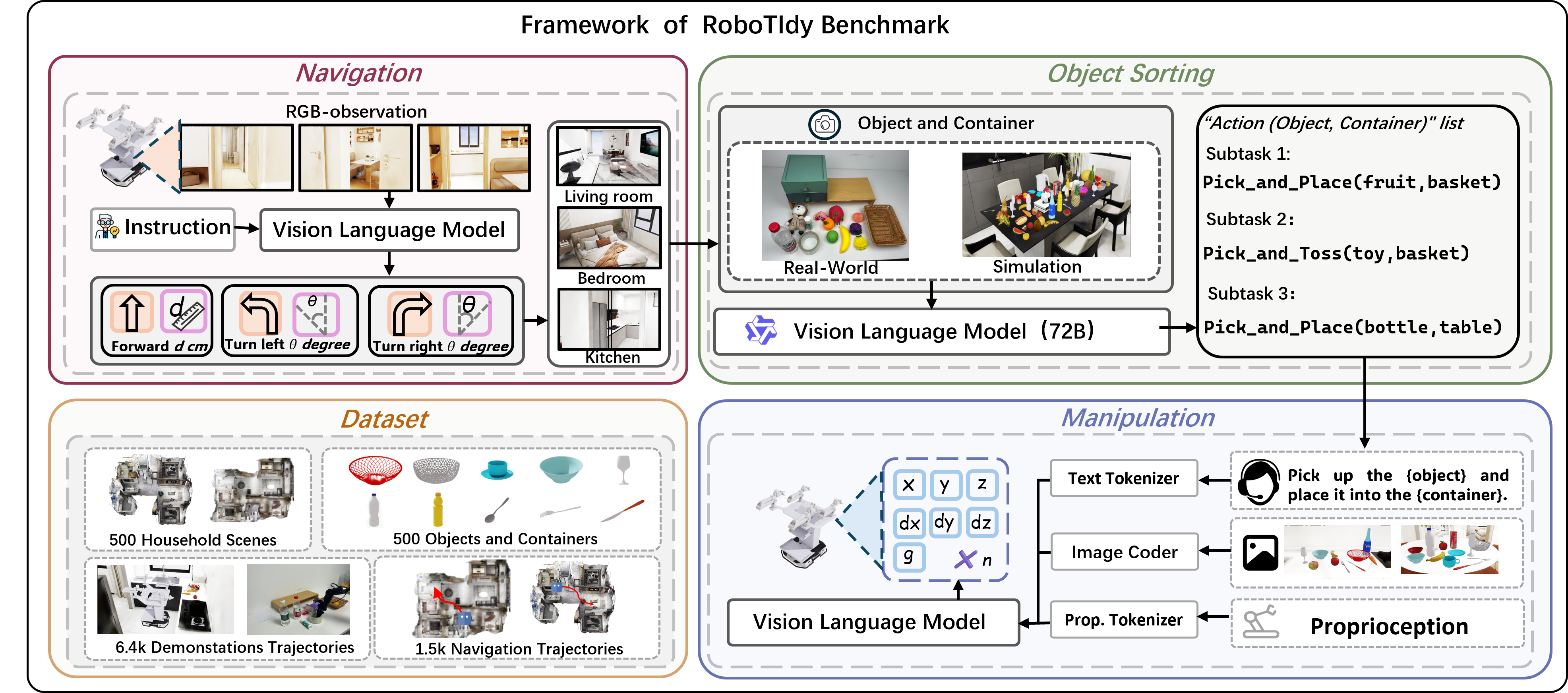}
    \captionsetup{font=footnotesize}
    \captionof{figure}{
    \textbf{Overview of RoboTidy Benchmark framework and dataset.} It spans Navigation, Object Sorting, and Manipulation: Qwen2.5-VL parses observations into an \textit{"Action (Object, Container)“ list} executed as manipulation actions. Our dataset offers 500 3DGS household scenes, 500 objects and containers, 6.4k manipulation trajectories and 1.5k navigation trajectories for sim2real evaluation.
    }
    \label{fig:teaser}
    \vspace{1.5em}
\end{center}
}]

\footnotetext[1]{Xiaoquan Sun and Ruijian Zhang contributed equally to this work.}
\footnotetext[2]{Corresponding author: Jiayu Chen, jiayuc@hku.hk}
\maketitle
\begin{abstract}
Household tidying is an important application area, yet current benchmarks neither model user preferences nor support mobility, and they generalize poorly, making it hard to comprehensively assess integrated language-to-action capabilities. To address this, we propose RoboTidy, a unified benchmark for language-guided household tidying that supports Vision-Language-Action (VLA) and Vision-Language-Navigation (VLN) training and evaluation. RoboTidy provides 500 photorealistic 3D Gaussian Splatting (3DGS) household scenes (covering 500 objects and containers) with collisions, formulates tidying as an "Action (Object, Container)" list, and supplies  6.4k high-quality manipulation demonstration trajectories and 1.5k naviagtion trajectories to support both few-shot and large-scale training. We also deploy RoboTidy in the real world for object tidying, establishing an end-to-end benchmark for household tidying. RoboTidy offers a scalable platform and bridges a key gap in embodied AI by enabling holistic and realistic evaluation of language-guided robots.
\end{abstract}    
\section{Introduction}
\label{sec:intro}

In recent years, household tidying has emerged as a prominent research avenue within indoor embodied intelligence. Fueled by rapid advances in Multimodal Large Language Models (MLLMs) and substantial gains in the capabilities of Vision-Language-Navigation (VLN)~\cite{VLNCE,NaVid,NaVILA} and Vision-Language-Action (VLA)~\cite{ACT,RDT,pi0.5}, an increasing number of studies now deploy and evaluate these models in real world environments to accomplish language-to-action tidying tasks in household scenes. 

% 在家居收纳（household tidying）中，核心任务是将每个对象放置到恰当的容器/位置。传统路线要么要求用户为每件物体逐一显式指定目标位置，过程繁琐且计算效率低（Rasch et al., 2019; Yan et al., 2021）；要么通过跨用户取平均学习非个性化的“通常放置”规则，忽视个体差异（Taniguchi et al., 2021; Kant et al., 2022; Sarch et al., 2022）。面向个性化的方法尝试基于少量示例外推用户偏好（如协同过滤、空间关系建模、潜在偏好向量等；Abdo et al., 2015; Kang et al., 2018; Kapelyukh and Johns, 2022），但往往依赖大规模偏好数据。与此同时，在真实环境中为 VLA/VLN 采集训练数据代价高且质量受限，研究社区因而普遍采用 sim-to-real 路线以获得大规模、可复现的数据。

In household tidying, the key task is to place every object into a suitable container. Traditional methods~\cite{tidy,yan2021} often require users to explicitly specify a target container for every object, a process that is cumbersome and computationally inefficient. Alternatively, they learn non-personalized “typical placement” rules by averaging across users, thereby ignoring individual differences~\cite{taniguchi,kant}. TidyBot~\cite{tidybot} infers from few-shot user preferences; it works in specific scenes but lacks robustness and generalization. To move beyond such per-user heuristics toward deployable policies, VLN and VLA models must be trained in environments with realistic visuals and physics.

However, collecting data in real-world environments for training VLA and VLN is costly and often of limited quality, prompting widespread adoption of the sim-to-real paradigm to obtain large-scale, high-quality datasets. To better bridge the sim-to-real gap, household scene representations have progressed from RGB-D–based scanned meshes such as Matterport3D~\cite{matterport3d} and HM3D~\cite{HM3D} to 3D Gaussian Splatting (3DGS)~\cite{3dgs}, which offers superior photorealism and cross-view consistency, thereby providing more stable visual conditions for policy learning in VLN and VLA.

We introduce RoboTidy, a unified benchmark for language-guided tidying that jointly supports VLA and VLN. We formalize tidying as an \textit{"Action (Object, Container)" list} with four manipulation actions and realize closed-loop execution in a physically realistic, photorealistic simulator; across-room navigation and multi-modal sensing are integrated for training and evaluation. On the data side, based on InteriorGS~\cite{InteriorGS}, we build 500 3DGS household scenes with convex-decomposed collision meshes and provide 6k manipulation demonstrations, 400 real world manipultion demonstrations and 1.5k navigation trajectories with language instructions.

Our experiments cover representative baselines for VLA and VLN. We observe that the VLN success rate on RoboTidy is markedly lower than that on R2R (Val-Unseen)~\cite{VLNCE}, underscoring the increased difficulty and realism of our benchmark; pretrained VLA models fine-tuned in our framework exhibit improved robustness in unseen scenes and with unseen objects, highlighting the role of diverse data and physics-aware simulation. We further validate sim2real transfer via real world mobile bimanual tidying experiments.
 
In summary, our key contributions are as follows.
\begin{itemize}    
    \item \textbf{Object Sorting}. We create an \textit{Action  (Object, Container) list}, and use Qwen2.5-VL to automatically infer placement rules from observations and decide the actions.    
    \item \textbf{3DGS household scenes}. We provide photorealistic 3DGS household scenes that cover 500 objects and containers. We also release 6.4k manipulation demonstration trajectories and 1.5k navigation trajectories to support training.
    \item \textbf{RoboTidy benchmark}. We evaluate representative VLA and VLN methods on our benchmark to assess generalization, and develop robot tidying in the real world to demonstrate sim2real transfer.
\end{itemize}

\section{Related work}
\label{sec:formatting}
%早期的家庭模拟基准通常通过要求机器人将物体移动至预定义目标位置来评估物品放置或重排任务（如 VirtualHome、ALFRED、TEACh、Habitat-2.0）。然而，这类方法高度依赖逐物体目标标注，难以支撑大规模训练与评测。近期的 TidyBot 虽利用大语言模型从少量文本示例中推理家庭收纳偏好，但其缺乏可扩展、物理真实的系统化评测框架，并在不同环境下表现出有限的鲁棒性。
Early household benchmarks~\cite{virtualhome, habitat, alfred,teach,lota-bench} typically evaluate object placement and rearrangement by requiring robots to move objects toward the pre-specified target position~\cite{housekeep,manipulathor,visualroom}. However, these methods rely heavily on per-object goal annotations, which make large-scale training and evaluation prohibitively expensive. More recently, TidyBot~\cite{tidybot} infers preferences via LLMs from few examples; however, it does not provide a scalable, physically realistic benchmark and shows limited robustness across varied environments.

Recent embodied AI benchmarks have explored household tasks from different perspectives, including semantic planning~\cite{virtualhome,alfred,teach}, multi-task manipulation~\cite{DBLP:conf/icml/ChenTLA23,arnold,robotwin,maniskill2}, and large-scale activity suites~\cite{Behavior-1k,colosseum}. However, these platforms mostly target generic task execution, offering neither a unified focus on tidying organization nor a joint evaluation of high-level reasoning and low-level, feedback-driven control under photorealistic, physically realistic conditions.

Our \textbf{RoboTidy} benchmark targets household tidying by combining photorealistic 3DGS household scenes with Isaac Sim 5.0~\cite{NVIDIA_Isaac_Sim}. It unifies manipulation and navigation through an ~\textit{"Action (Object, Container)" list} and navigation trajectories, supporting both VLA~\cite{ACT,RDT,pi0.5} and VLN~\cite{NaVid,NaVILA,VLNCE}. RoboTidy enables modular and feedback-aware evaluation with built-in trajectory logging and photorealistic execution.

\definecolor{BrightGreen}{HTML}{34C759} % 清亮绿
\definecolor{BrightRed}{HTML}{FF3B30}   % 系统红风格
% 勾/叉（不依赖 pifont）
\DeclareRobustCommand{\cmark}{\textcolor{BrightGreen}{\usym{2713}}}
\DeclareRobustCommand{\xmark}{\textcolor{BrightRed}{\usym{2717}}}
% \newcommand{\cmark}{\textcolor{BrightGreen}{\checkmark}}
% \newcommand{\xmark}{\textcolor{BrightGreen}{\checkmark}}

% ====== 表格 ======
\begin{table}[h]
\centering
\scriptsize
\setlength{\tabcolsep}{4pt}
\renewcommand{\arraystretch}{1.15}
\caption{Comparison of various frameworks and benchmarks. \cmark\ indicates presence; \xmark\ indicates absence; ``--'' indicates not reported.}
\begin{tabular}{l*{11}{c}}
\toprule
\textbf{Feature} &
\rotatebox{90}{\textbf{RoboTidy}} &
\rotatebox{90}{\textbf{tidybot~\cite{tidybot}}} &
\rotatebox{90}{\textbf{Habitat 2.0~\cite{habitat}}} &
\rotatebox{90}{\textbf{arnold~\cite{arnold}}} &
\rotatebox{90}{\textbf{RLBench~\cite{rlbench}}} &
\rotatebox{90}{\textbf{Behavior-1K~\cite{Behavior-1k}}} &
\rotatebox{90}{\textbf{Teach~\cite{teach}}} &
\rotatebox{90}{\textbf{ManiSkill 2~\cite{maniskill2}}} &
\rotatebox{90}{\textbf{LIBERO~\cite{libero}}} &
\rotatebox{90}{\textbf{CALVIN~\cite{calvin}}} \\
\midrule
Manipulation & \cmark & \cmark & \cmark & \cmark & \xmark & \cmark & \cmark & \cmark  & \cmark & \cmark \\
Navigation  & \cmark & \cmark & \cmark & \xmark & \xmark & \cmark & \cmark & \cmark  & \xmark & \xmark \\
Household Scenes          & \cmark & \cmark & \cmark & \cmark & \xmark & \cmark & \cmark & \xmark & \xmark & \xmark \\
Realistic Object Physics   & \cmark & \xmark & \xmark & \cmark & \cmark & \cmark & \cmark & \xmark  & \cmark & \xmark \\
Photorealism & \cmark & \cmark & \cmark & \xmark & \xmark & \cmark & \xmark & \cmark  & \cmark & \xmark \\

\midrule
Num Scenes & 500    & 96     & 1      & 15     & 1      & 50     & 3      & --   & 20     & 1 \\
\bottomrule
\end{tabular}
\label{tab:benchmark-compare}

\end{table}

\section{Method}
\subsection{Overview}
%我们将家庭场景收纳任务表述为：在家庭多房间环境中对物品进行有序归纳，自动生成并维护“物品—容器—动作原语”清单，同时执行跨房间的收纳移动，并全程采集可直接支持 VLA 与 VLN 训练和评测的数据。In this paper,我们提出 RoboTidy：一个面向家具场景的收纳—导航、联合评测的基准。方法上，我们在具备物理真实性与照片级真实感( physically and photorealistic)的 NVIDIA Isaac Sim 上搭建模块化框架，系统以多视角观测为输入，利用 VLM 提取物品和容器的类别与属性，对新物品直接决定目标容器与最小四原语（抓取-放置、抓取-投掷、开启、关闭）。在此基础上，RoboTidy 包含三个关键模块：1)物品收纳模块按规则将物品放置至指定容器并更新清单；2)动作原语模块:基于 IK 求解器生成可行轨迹实施原语并记录视觉-动作数据，用于 VLA训练和测评；3)移动模块:依托全局地图进行路径规划与跨房间导航，同步生成导航轨迹以支持 VLN训练和测评。4

We formulate our problem as performing structured organization of objects in household scenes, automatically generating an \textit{"Action (Object, Container)" list}, executing cross-room tidying, and continuously logging data that supports VLA and VLN training and evaluation. In this paper, we present RoboTidy, a unified benchmark for
language-guided tidying that jointly supports VLA and
VLN for household scenes. Methodologically, we build a modular framework in NVIDIA Isaac Sim~\cite{NVIDIA_Isaac_Sim} with physical and photorealistic simulation: the system takes multi-view observations as input, uses Qwen2.5-VL~\cite{qwen2.5VL} to extract category and attribute semantics of objects and containers, and for newly encountered items directly selects a target container and one of four manipulation actions \textit{(Pick and Place, Pick and Toss, Open the Container, Close the Container)}.

Our benchmark, RoboTidy, consists of four key modules detailed in the following sections: a) Tidying module that places objects into target containers according to the induced rules and updates the list; b) Manipulation action module that employs an inverse kinematics (IK) controller to generate feasible trajectories, executes the action, and logs vision–action data for VLA training and evaluation; c) Navigation module that performs path planning and across-room navigation while recording navigation trajectories to support VLN training and evaluation.; d) Sensor module that supports multimodal RGB-D and LiDAR sensing and handles data collection in simulation.
\subsection{Object Sorting}
%我们提出一种无需额外示例输入的偏好归纳方法：对当前操作台上面的容器以及容器内的物品进行扫描，接读取现有的“容器→物体”清单。随后，利用视觉语言模型读取物体语义与属性信息，并将观测自动转写为可解析的代码式提示。在此基础上，LLM 对“物体—容器”对应关系进行概括与归纳，生成个性化收纳规则（如“水果→篮子，衣物→柜子”），并对桌面上未进行归纳的物品时给出“容器”与“动作”的决策。we designed 4 actions. The actions include:pick and place, pick and toss, open the drawer, close the drawer
\begin{figure}[t]  
    \centering
    \includegraphics[width=0.95\linewidth]{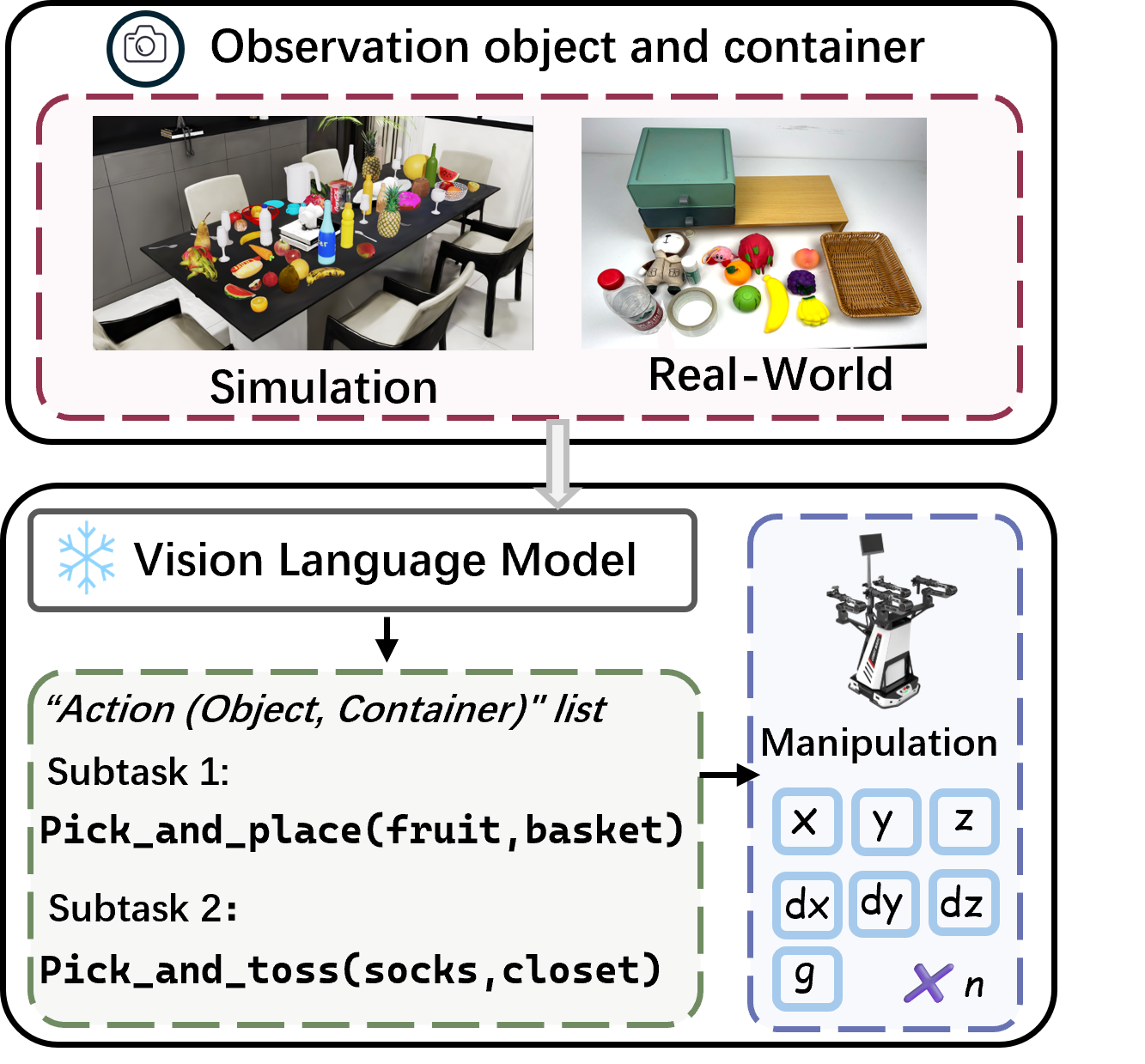} 
    \caption{\textbf{Object Sorting Pipline.} From workspace observations, Qwen2.5-VL~\cite{qwen2.5VL} identifies objects and containers and produces an \textit{Action (Object, Container) list}, which the system executes manipulation actions to complete sorting.}
    \label{fig:mobile_task} 
\end{figure}
We propose a preference tidying method that does not require additional input. At first the system passively scans the containers on the current workspace and their contents to obtain the existing "Object →  Container". Leveraging a Vision Language Model (VLM), it extracts object semantics and attributes, and automatically converts these observations into a parsable, code-style prompt. On this basis, an LLM abstracts the observed object → container correspondences into a personalized object sorting \textit{"Action (Object, Container)" list} and for newly observed items that are not yet covered by the induced rules, jointly decides both the target container and the action to execute. We design a minimal set of four manipulation actions:
\begin{itemize}
    \item \textit{Pick and Place}: Use the gripper to grasp the object at its detected center, move the gripper to a pose just above the selected receptacle, and place the object at the target position.
    \item \textit{Pick and Toss}: Use the gripper to grasp the object at its detected center, then swing the arm and release the gripper with appropriate timing to toss the grasped object into the selected receptacle~\cite{tossingbot}.
    \item \textit{Open the Container}: Switch the container (e.g., drawer / cabinet) from closed to open.
    \item \textit{Close the Container}: Switch the container(e.g., drawer / cabinet) from open to closed.
\end{itemize}
To account for diverse household tidying preferences, we annotate for each object the set of acceptable containers and, when multiple options exist, a priority order specifying the primary container and its equivalents, using the following criteria:
\begin{itemize}
\item \textbf{Attribute:} Sort objects by observable attributes such as material or size (e.g., put plastic objects in one container and metal ones in a different container).
\item\textbf{Function:} Sort objects by purpose or usage context (e.g., put clothes here and socks there).
\item\textbf{Safety:} Sort objects by safety risk and fragility (e.g., keep sharp tools in a closed container).
\item\textbf{Hygiene:} Sort objects by cleanliness and contamination risk (e.g., put waste into a trash bin and place clean clothes in a wardrobe).
\end{itemize}
When multiple criteria are applied, the primary container is chosen according to the preset priority and the remaining options are included in the correct set as equivalent containers. All annotations are frozen before evaluation.

\subsection{Navigation Module}

%The navigation module consists a path planner and a low-level controller.The path planner利用场景信息和导航要求生成速度相关的运动指令，The controller则根据运动指令控制机器人动作。我们考虑了两种不同条件下的顶层规划策略：
%全局地图已知。我们生成了诸多参考轨迹并向一个PID控制器输入上述轨迹离散点和机器人的实时位置以减小机器人跟踪误差。
%全局地图未知。进一步地，我们研究了仅依赖视觉信息与自然语言指令完成导航的情况。根据...多模态信息，VLM会直接输出速度相关的运动指令指导机器人运动。(有些多余，后面基本都是废话，就是一句话就可以说明白的)
%我们利用强化学习PPO算法完成对智能体运动姿态的底层控制。PPO策略网络接收上层传来的速度指令并输出作用到机器人各关节的电机指令，完成对机器人物理层面的驱动。

%两个问题：
%Measurement(?不确定要不要写在这里，即navigation相关的指标)
%前面叙述部分的详略调整，措辞调整，内容得当调整等等，是否应该加入一些基本的公式性质的原理解释？（感觉有些累赘）
The navigation module comprises a path planner and a low-level controller. During data generation, we run the A* planner~\cite{A*} on the InteriorGS~\cite{InteriorGS} 2D semantic map to plan reference paths, time-parameterize and discretize them into waypoints; a PID controller tracks the trajectory using the robot’s real time pose, outputs velocity commands, and logs full navigation trajectories for VLN training and evaluation. During VLM evaluation, no map is supplied camera images and language instructions form a multimodal input that drives the VLM to directly predict velocity commands for navigation. For low-level control, a Proximal Policy Optimization (PPO)~\cite{PPO} policy converts velocity commands into joint-level motor actions to drive the robot.

\subsection{Manipulation Module}
%我们使用 IK 求解器与运动规划器完成动作原语的轨迹规划与执行，并采集示例轨迹用于 VLA的训练和evaluation。首先给定物体(object) 6D 位姿与目标容器的位置，系统自动生成预抓取、抓取与预放置、放置的动作；由 IK求解器选择候选关节路点，结合关节限位、速度、加速度与环境碰撞约束进行避碰。，夹爪以开合宽度阈值闭环判定抓取与释放。我们同步记录多视角RGB图像、关节位姿与夹爪(Gripper)状态以及对应的动作原语，生成可直接用于 VLA 训练和评测的轨迹。
We use an inverse kinematics (IK) solver together with a motion planner to plan and execute trajectories for the manipulation actions, and we collect demonstration trajectories for VLA training and evaluation. Given an object’s 6D pose and the position of the target container, the system automatically generates the pre-grasp, grasp, pre-place, and place; the IK solver proposes candidate joint waypoints, and the motion planner computes collision-free trajectories under joint limits, velocity, acceleration, and environment collision constraints. The gripper employs a closed-loop width-threshold criterion to determine grasp and release. We synchronously log multi-view RGB images, joint configurations and gripper states, along with the corresponding action-primitive labels, producing trajectories directly usable for VLA training and evaluation.

\subsection{Sensors Module}
Our benchmark supports multimodal data collection using RGB-D cameras and LiDAR scanning.  In the simulation environment, new sensors are integrated by mounting them on the robot or within the scene, configuring their parameters, and referencing their prim paths in the main configuration. In the real-world environment, the number of cameras and LiDAR units is constrained by compute resources, I/O throughput, and interface bandwidth.

 \subsection{Metrics}
\textbf{Object Placement Accuracy (OPA).} We treat placement as the selection from a scene’s closed set of candidate containers. 
For a scene $s$ and the set of all scenes $S$, the per scene \(\mathrm{OPA}(s)\) and overall OPA are:
\[
\mathrm{OPA}(s)=\frac{C(s)}{N(s)},\qquad
\mathrm{OPA}=\frac{1}{|S|}\sum_{s\in S}\mathrm{OPA}(s).
\]
\noindent\textbf{Valid Sorting Success Rate (VSSR).}
An object counts as a success if its predicted container is valid (same rule as OPA)
and its required actions finish successfully.

For a scene $s$, let $v_j(s)\in\{0,1\}$ indicate success for object $j$. Each scene $\mathrm{VSSR}(s)$ and the overall VSSR are:
\[
\mathrm{VSSR}(s)=\frac{1}{N(s)}\sum_{j=1}^{N(s)} v_j(s),\quad
\mathrm{VSSR}=\frac{1}{|S|}\sum_{s\in S}\mathrm{VSSR}(s).
\]
where $s$ denotes a household scene, $S$ denotes the set of all household scenes, $N(s)$ denotes the total number of objects in $s$, and $C(s)$ denotes the number of objects whose predicted container is judged correct. Predictions that fail to parse as a valid container, or that are not in the scene’s candidate container set, are counted as incorrect. For \textbf{VSSR}, $v_j(s)=1$ iff $c_j(s)=1$ and $d_j(s)=1$; $c_j(s)=1$ denotes a valid container prediction (predictions outside the candidate set count as $0$), and $d_j(s)=1$ denotes completion of the required manipulation actions (failures or aborts count as $0$).

%additional experimental details can be found in Supplementary Material.
\section{Experiment}
\subsection{Benchmark Dataset}
%在实验部分，我们基于 InteriorGS 的 3DGS 资产构建了包含 675 个家庭场景的数据集，覆盖 living room、bedroom、kitchen 三类房间。每个场景包含一组可操作物品与 2–5 个收纳容器（如篮子、书架、衣柜等）。我们按“每容器 2 条已见 + 2 条未见”构造〈物品→容器〉放置记录，因此每景可提供 4–10 条场内已见示例（仅用于 FS 设定）。我们评估两种放置预测任务：Zero-Shot（ZS）——仅给定场景观测，不提供任何场内示例，模型需为未见物品预测目标容器；Few-Shot（FS）——在相同场景上额外提供上述已见示例（每容器 2 条，共 4–10 条）以归纳偏好并预测其余未见物品。数据集总体规模为 300 个物品实例与 20 个容器类别。为确保可复现与公平对比，我们对所有方法使用共享且固定的已见/未见划分与随机种子，并以 Top-1 准确率为主指标（按场景与按容器进行 macro-average）。
\noindent\textbf{3DGS household scenes.} Building on InteriorGS~\cite{InteriorGS}, We construct 500  household scenes across \textit{living room}, \textit{bedroom}, and \textit{kitchen}. 
Although 3DGS assets are appearance-only, each scene is authored as a 3DGS--mesh hybrid~\cite{miao}: 3DGS (USDZ via 3DGUT~\cite{wu20253dgut}) remains visible for high-fidelity rendering, and artist-created triangle meshes are convex-decomposed with CoACD~\cite{Wei2022CoACD} to produce per-object collision bodies. We assemble USDA scenes with these collision shapes authored as invisible rigid bodies, enabling accurate contact and dynamics in Isaac~Sim~5.0~\cite{NVIDIA_Isaac_Sim}.

\noindent\textbf{Object Assets.} We create an extensive collection of 3D assets for daily items to meet diverse scene construction needs. This collection covers categories such as fruits, beverages, containers, tableware and decorations. These assets are sourced from online 3D model repositories Sketchfab and TurboSquid, and have undergone standardized format processing. Our screening process involves removing low-precision models and those with appearance defects to ensure the high-quality presentation of the assets.

\noindent\textbf{Demonstration Trajectories.} We collected demonstration trajectories for four manipulation actions. For each action, data were gathered in two household scenes, covering five object–container categories and three rooms (\textit{kitchen}, \textit{bedroom}, \textit{living room}). For every "category–scene–room" combination, we recorded 100 demonstration trajectories, yielding 6k manipulation trajectories in total.
For navigation, we fuse an occupancy grid with a 2D semantic map to construct a navigation map that is both traversability-aware and semantically grounded, and run an A*-based planner~\cite{A*} to generate paths. Across 500 household scenes, we instantiate one trajectory for each pair among the three rooms, resulting in 1.5k navigation trajectories.
%We collected  demonstrations trajctories across four actions: for each action, under two household scenes, spanning five object and container categories and three room \textit{(kitchen, bedroom and livingroom)}. For every category–setting–room combination, we gathered 100 manipulation demonstration trajectories, yielding 6k trajectories in total. To construct navigation trajectories, we first fuse the occupancy grid map with a 2D semantic map to obtain a navigation map that is both traversability-aware and semantically interpretable. We then run an A*-based planner on this map to generate trajectories. In every household scene, we consider all pairwise combinations among three room types (\textit{kitchen}, living room, and bedroom) and generate one trajectory for each pair. Across 500 independent scenes, this yields a total of 1,500 trajectories.
%为构建导航轨迹，我们首先将占据栅格地图与二维语义地图进行融合，得到兼具可通行性与语义可解释性的导航底图；随后在该底图上运行基于 A* 的算法，生成跨房间的导航轨迹。，在每个household场景中从三类房间kitchen、living room、bedroom）中选取所有两两组合，并为每一组生成一条导航轨迹，。在 500 个独立场景下，总计获得 1,500 条导航轨迹。

% 语言指令。对于导航而言，我们综合考虑以下方面人工完成对高层指令的设计：语义贴合现实需求，增强导航的合理性；将场景中具有鲜明视觉特征的物品作为“地标”，并通过相对空间位置关系细化引导描述。底层指令则由“前进”、“转向”等基本移动方式构成。 最终，我们在x个场景中构建了x条语言指令.
%操纵部分。 采用基于模板的指令，绑定四类动作元（pr，并以 对象、容器 及可选属性为插槽；通过受控词汇变体增强多样性且保持语义一致（如：“把苹果放进篮子里”）。导航部分。 采用分层方案：高层指令遵循两项原则——(i) 语义与真实意图对齐；(ii) 基于地标并通过相对空间关系细化引导；底层指令由基本运动原语组成（如“前进”“左/右转”）。针对每条路径，我们生成一条高层指令，并可选配套一段底层步骤。
\noindent\textbf{Language Instructions.} We employ template-based commands for four manipulation actions with slots for the object and container; controlled lexical variants increase diversity while preserving semantics (e.g., \textit{“pick up the [object] and place into the [container]”}, \textit{“pick up the [object] and toss into [container]”}). For navigation, For navigation, we adopt a hierarchical method~\cite{miao}. High-level instructions align with realistic intents and are landmark-grounded via relative spatial relations; low-level instructions comprise basic actions \textit{(e.g., move forward, turn left or right)}; for each path, an LLM produces one high-level instruction and an associated sequence of low-level action steps as the final instruction.

\begin{table*}[t]
\centering
\small
\caption{Comparsion among of  diffent object sorting methods. We report \textbf{OPA} (\%) for each method under \textbf{ZS} and \textbf{FS} setting, with  mean and standard deviation over 3 random seeds.}
\label{tab:object_sorting}
\setlength{\tabcolsep}{6pt}           
\renewcommand{\arraystretch}{1.05}    
\begin{tabular}{@{}l cc cc cc cc@{}}
\toprule
\multirow{2}{*}{Room} 
& \multicolumn{2}{c}{RoBERTa~\cite{Roberta}} 
& \multicolumn{2}{c}{CLIP~\cite{CLIP}} 
& \multicolumn{2}{c}{TidyBot~\cite{tidybot}} 
& \multicolumn{2}{c}{RoboTidy (ours)} \\
& ZS  & FS  
& ZS  & FS 
& ZS  & FS 
& ZS  & FS  \\
\midrule
Livingroom  &
71.5{\color{gray!80}\footnotesize$\,\pm\,1.2$}&
78.4{\color{gray!80}\footnotesize$\,\pm\,1.7$}&
79.6{\color{gray!80}\footnotesize$\,\pm\,2.6$}&
84.5{\color{gray!80}\footnotesize$\,\pm\,1.9$}&
94.5{\color{gray!80}\footnotesize$\,\pm\,0.7$}&
95.1{\color{gray!80}\footnotesize$\,\pm\,2.4$}&
89.5{\color{gray!80}\footnotesize$\,\pm\,0.6$}&  
92.4{\color{gray!80}\footnotesize$\,\pm\,2.0$} \\
Kitchen &
77.3{\color{gray!80}\footnotesize$\,\pm\,0.7$}&
84.5{\color{gray!80}\footnotesize$\,\pm\,0.9$}&
83.1{\color{gray!80}\footnotesize$\,\pm\,1.8$}&
87.3{\color{gray!80}\footnotesize$\,\pm\,2.0$}&
88.5{\color{gray!80}\footnotesize$\,\pm\,1.6$}&
90.1{\color{gray!80}\footnotesize$\,\pm\,0.9$}&
96.3{\color{gray!80}\footnotesize$\,\pm\,2.8$}&
97.1{\color{gray!80}\footnotesize$\,\pm\,1.4$}\\
Bedroom  &
76.3{\color{gray!80}\footnotesize$\,\pm\,0.9$}&
79.1{\color{gray!80}\footnotesize$\,\pm\,1.6$}&
81.2{\color{gray!80}\footnotesize$\,\pm\,1.3$}&
85.3{\color{gray!80}\footnotesize$\,\pm\,2.5$}&
89.3{\color{gray!80}\footnotesize$\,\pm\,1.7$}&
71.9{\color{gray!80}\footnotesize$\,\pm\,0.6$}&
92.3{\color{gray!80}\footnotesize$\,\pm\,1.2$}&
94.1{\color{gray!80}\footnotesize$\,\pm\,1.3$}\\
\cmidrule(lr){1-9}
Average OPA &
75.1{\color{gray!80}\footnotesize$\,\pm\,0.7$} & 80.6{\color{gray!80}\footnotesize$\,\pm\,1.2$} & 81.9{\color{gray!80}\footnotesize$\,\pm\,1.8$} & 85.6{\color{gray!80}\footnotesize$\,\pm\,2.1$} & 
90.7{\color{gray!80}\footnotesize$\,\pm\,1.4$} & 85.7{\color{gray!80}\footnotesize$\,\pm\,1.8$} & 92.7{\color{gray!80}\footnotesize$\,\pm\,1.9$} & 94.6{\color{gray!80}\footnotesize$\,\pm\,1.7$} \\
\bottomrule
\end{tabular}
\end{table*}

\begin{table*}[h]
  \centering
  \small
  \setlength{\tabcolsep}{3pt}
  \caption{Comparison of VLN methods on the RoboTidy benchmark. We report \textbf{SR}, \textbf{OSR}, and \textbf{SPL} for each method, with mean ± standard deviation over 3 random seeds.}
  \begin{tabular}{lccccccc}
    \toprule
    \textbf{Metric} & CMA~\cite{VLNCE} & InternVL-2.5-8B~\cite{InternVL} & Llama-3.2-11B~\cite{Llama3} & NaVid-base~\cite{NaVid} & NaVILA-base~\cite{NaVILA} & Navid-R (ours) & NaVILA-R (ours) \\
    \midrule
    SR   & 
0.09{\color{gray!80}\footnotesize$\,\pm\,0.01$} &
0.08{\color{gray!80}\footnotesize$\,\pm\,0.01$} & 
0.12{\color{gray!80}\footnotesize$\,\pm\,0.01$} & 
0.11{\color{gray!80}\footnotesize$\,\pm\,0.02$} & 
0.22{\color{gray!80}\footnotesize$\,\pm\,0.04$} & 
0.16{\color{gray!80}\footnotesize$\,\pm\,0.01$} &
0.26{\color{gray!80}\footnotesize$\,\pm\,0.01$} \\
    OSR  & 
0.11{\color{gray!80}\footnotesize$\,\pm\,0.01$} & 
0.11{\color{gray!80}\footnotesize$\,\pm\,0.02$} & 
0.16{\color{gray!80}\footnotesize$\,\pm\,0.03$} & 
0.12{\color{gray!80}\footnotesize$\,\pm\,0.00$} & 
0.12{\color{gray!80}\footnotesize$\,\pm\,0.00$} &
0.14{\color{gray!80}\footnotesize$\,\pm\,0.01$} &
0.15{\color{gray!80}\footnotesize$\,\pm\,0.02$} \\
    SPL  & 
0.10{\color{gray!80}\footnotesize$\,\pm\,0.03$} & 
0.12{\color{gray!80}\footnotesize$\,\pm\,0.01$} & 
0.13{\color{gray!80}\footnotesize$\,\pm\,0.00$} & 
0.11{\color{gray!80}\footnotesize$\,\pm\,0.01$} &
0.20{\color{gray!80}\footnotesize$\,\pm\,0.02$} &
0.15{\color{gray!80}\footnotesize$\,\pm\,0.00$} &
0.23{\color{gray!80}\footnotesize$\,\pm\,0.02$}\\
    \bottomrule
  \end{tabular}
  \label{tab:vln}
\end{table*}

\begin{table}[t]
\centering
\small
\caption{Comparison among different sizes of Qwen2.5-VL~\cite{qwen2.5VL} in terms of \textbf{OPA} (\%) under both \textbf{ZS} and \textbf{FS} settings. Results are reported as mean ± standard deviation over 3 random seeds. Here, “B” denotes billions.}
\begin{tabular}{lccc}
\toprule
VLM Model & params & ZS & FS \\
\midrule
Qwen2.5-VL       &0.5B    
& 92.6{\color{gray!80}\footnotesize$\,\pm\,1.7$} 
& 94.9{\color{gray!80}\footnotesize$\,\pm\,3.2$}  \\

Qwen2.5-VL       &7B      
& 93.7{\color{gray!80}\footnotesize$\,\pm\,3.1$}
& 95.2{\color{gray!80}\footnotesize$\,\pm\,3.6$}  \\

Qwen2.5-VL       & 72B (ours)    
& 94.0{\color{gray!80}\footnotesize$\,\pm\,2.9$}
& 95.6{\color{gray!80}\footnotesize$\,\pm\,2.6$}  \\
\bottomrule
\end{tabular}

\label{tab:qwen2.5VL}
\end{table}

\subsection{Baselines}
We benchmark across three task types: object sorting, manipulation, and navigation. For the object sorting task, we test RoBERTa~\cite{Roberta}, CLIP~\cite{CLIP}, and TidyBot~\cite{tidybot}. For the manipulation task, we adopt ACT~\cite{ACT}, RDT~\cite{RDT}, and 
\(\pi_{0.5}\)~\cite{pi0.5}. For the navigation task, we adopt VLN-CE~\cite{VLNCE}, NaVid~\cite{NaVid} and NaVILA~\cite{NaVILA}.

\noindent\textbf{Object Sorting Baseline.}
\begin{itemize}
    \item \textbf{RoBERTa~\cite{Roberta}:} A Sentence-BERT~\cite{sentence} name embeddings map each unseen object to the receptacle of its cosine-nearest seen neighbor.

    \item \textbf{CLIP~\cite{CLIP}:} A CLIP-based name-embedding method that embeds object names with CLIP’s text encoder and maps each unseen object to the receptacle of its nearest seen neighbor in the shared semantic space using cosine similarity.

    \item \textbf{TidyBot~\cite{tidybot}:} A few-shot preference–summarization approach that uses an LLM to infer user-specific  rules from textual examples and grounds the summarized categories with open-vocabulary image classification.
\end{itemize}
\noindent\textbf{Manipulation Baseline.}
\begin{itemize}
    \item \textbf{ACT~\cite{ACT}:} A CVAE-based imitation learner that chunks actions to predict future action sequences and uses temporal ensembling for smooth, stable execution.
    \item \textbf{RDT~\cite{RDT}:} A diffusion-augmented transformer~\cite{RDT} for bimanual manipulation that unifies the action space and fuses multimodal inputs, enabling data-efficient few-shot learning across diverse tasks.
    \item \textbf{$\pi_{0.5}$~\cite{pi0.5}:} A scalable VLA policy with semantic planning and flow-matching control for open-world long-horizon mobile manipulation in unseen homes.
\end{itemize}
\noindent\textbf{Navigation Baseline.}
\begin{itemize}
    % 一个以指令为导向的导航基准：在连续、照片级真实的 3D 环境中，采用众包的语言指令，智能体可在连续动作空间中自由导航。(VLNCE是一种方法，这里不能说是一种benchmark，他的主语应该是method这种并非benchmark)
    \item \textbf{VLN-CE~\cite{VLNCE}:} An instruction-guided navigation benchmark in continuous, photo-realistic 3D environments with crowdsourced language and unconstrained agent control.
    % 一种基于视频的大型视觉语言模型，在无需地图、里程计或深度信息的情况下，将实时单目RGB视频流映射为下一步动作，并编码历史时空信息以支持指令跟随。
    \item \textbf{NaVid~\cite{NaVid}:} A video-based large VLM that maps on-the-fly monocular RGB streams to next-step actions without maps/odometry/depth, encoding spatio-temporal history for instruction following.
    % 一种通用型的具身导航模型：通过基于模式（schema-based）的指令来适配大语言模型（LLM），将多样化任务统一表述为文本生成问题。
    \item \textbf{NaVILA~\cite{NaVILA}:} A two-level framework that couples VLA planning with locomotion skills for navigation—issuing high-level language commands while a real-time locomotion policy handles obstacle avoidance.

\end{itemize}

\subsection{Object Sorting Task}
\textbf{Setting details.} Each scene contains a set of objects and 2–5 containers. We evaluate two object sorting settings: Zero-Shot (\textbf{ZS}), in which the model receives only scene observations without examples and must predict the target container for unseen objects, and Few-Shot (\textbf{FS}), in which the same scene is augmented with the seen examples (two per container, 4–10 per scene) to induce preferences and predict the remaining unseen objects. We use Qwen2.5-VL-72B~\cite{qwen2.5VL} as Vision Language Model backbone.
%Tabel 2显示，我们所提出的收纳方法在 Zero-Shot (ZS) 与 Few-Shot (FS) 设置下均表现出显著优势，整体性能明显优于所有基线方法。这种优势得益于我们方法采用了强大的视觉语言模型（VLM）通过对多模态信息的理解并从已见示例中抽象出通用整理规则，即使在没有任务示例的情况下，也能准确预测未见物体的目标容器，展现出极强的泛化能力。相比之下，RoBERTa 和 CLIP 主要依赖文本或单模态特征，难以捕捉家庭收纳中的语义与感知关联；而 tidybot 虽在特定的场景下表现良好，但在更复杂、干扰更多的环境中鲁棒性不足。

\noindent\textbf{Evaluation results.} \Cref{tab:object_sorting} shows that our method achieves state-of-the-art performance under both \textbf{ZS} and \textbf{FS} settings, significantly outperforming all baselines~\cite{Roberta,CLIP,tidybot}. This stems from our use of a powerful VLM~\cite{qwen2.5VL}, which abstracts object sorting rules through multimodal understanding even without demonstrations, it accurately predicts containers for unseen objects. In contrast, RoBERTa~\cite{Roberta} and CLIP~\cite{CLIP} rely on single modal features and struggle to model the semantic-perceptual relationships in the scene, while TidyBot~\cite{tidybot}, though effective in specific scene, lacks robustness in complex environments. As shown in \Cref{tab:qwen2.5VL}, we compare Qwen2.5-VL~\cite{qwen2.5VL} with different parameter sizes and find that while the \textbf{OPA} metric improves with increasing model size, the improvement is not linear.

\subsection{Manipulation Task}
%为了评估RoboTdiy的基准效用和泛化挑战，我们评估了三种策略模型：ACT、RDT、Pi0.5。所有的 VLA 都是在其发布的预训练权重基础上进行单个动作微调。评估在4个任务上进行，使用 Aloha AgileX 双臂机器人进行实验。对于每个任务，使用 100 条的示范轨迹进行训练，测试的时候我们在未见场景与未见物体和容器实例上进行测试，以衡量模型对环境与外观变化的鲁棒性。
\noindent\textbf{Setting details.} To assess the utility and generalization challenge of the RoboTidy benchmark, we compare three policies: ACT~\cite{ACT}, RDT~\cite{RDT}, and~\(\pi_{0.5}\)\cite{pi0.5}. All VLA methods are fine-tuned per primitive starting from their released pretrained weights. Experiments are conducted on four manipulation tasks using an aloha-agilex robot. For each task, we train on 100 demonstration trajectories and evaluate on unseen household scenes, objects and containers to measure robustness to environment. Additionally, under these unseen scenes, objects and containers conditions, we run 100 evaluation trials per action task and report the success rate (\%); Object sorting correctness is not assessed; we only measure whether the action task completes successfully.

\begin{table}[t]
\centering
\small
\caption{ Comparison among different VLA methods on our RoboTidy benchmark and report the success rate (\%) ± standard deviation over 3 random seeds.}
\begin{tabular}{lccc}
\toprule
Action Task & ACT~\cite{ACT} & RDT~\cite{RDT} & \(\pi_{0.5}~\cite{pi0.5}\) \\
\midrule
\textit{Pick and Place}       
& 24.2{\color{gray!80}\footnotesize$\,\pm\,1.2$}    
& 95.1{\color{gray!80}\footnotesize$\,\pm\,2.9$}
& 98.0{\color{gray!80}\footnotesize$\,\pm\,0.0$}  \\
\textit{Pick and Toss}      
& 6.2{\color{gray!80}\footnotesize$\,\pm\, 2.2$}      
& 73.7{\color{gray!80}\footnotesize$\,\pm\,2.0$}
& 85.2{\color{gray!80}\footnotesize$\,\pm\,1.4$}  \\
\textit{Open the Container}       
& 14.0{\color{gray!80}\footnotesize$\,\pm\,1.0$}    
& 83.2{\color{gray!80}\footnotesize$\,\pm\,2.0$} 
& 93.1{\color{gray!80}\footnotesize$\,\pm\,2.9$}  \\
\textit{Close the Container}       
& 21.3{\color{gray!80}\footnotesize$\,\pm\,0.8$}    
& 85.9{\color{gray!80}\footnotesize$\,\pm\,1.2$} 
& 96.2{\color{gray!80}\footnotesize$\,\pm\,2.6$}  \\
\bottomrule
\end{tabular}

\label{tab:VLA}
\end{table}

\begin{table}[t]
  \centering
  \small
  \setlength{\tabcolsep}{7pt}
  \caption{Comparison among different VLN methods on VLN-CE  benchmark~\cite{VLNCE} under \textbf{R2R Val-Unseen} setting. Results are reported as \textbf{SR}, \textbf{OSR} and \textbf{SPL}.}
  \begin{tabular}{lccc}
    \toprule
    \multirow{2}{*}{Methods} & \multicolumn{3}{c}{R2R Val-Unseen} \\
    \cmidrule(lr){2-4}
    & SR  & OSR  & SPL  \\
    \midrule
    Seq2Seq~\cite{seq2seq} & 0.25 & 0.37 & 0.22\\
    InstructNav~\cite{instructnav} & 0.42 & 0.29 & 0.12 \\
    Navid-base~\cite{NaVid}  & 0.22 & 0.32 & 0.17 \\
    NaVILA-base~\cite{NaVILA}  & 0.29 & 0.38 & 0.27 \\
    Navid-R (ours)  & 0.27 & 0.34 & 0.19 \\
    NaVILA-R (ours) & 0.31 & 0.41 & 0.28 \\
    \bottomrule
  \end{tabular}
  \label{tab:vlnce_results}
\end{table}
\noindent\textbf{Evaluation results.}
%表4展示了四类动作的评测结果。未预训练的 ACT 在未见场景与未见对象上表现不佳；相比之下，采用预训练的 RDT 与 Pi0.5 在不同的家具场景下保持更强的稳定性，表明视觉–语言–动作（VLA）预训练为跨域泛化提供了有效先验。在我们的评测设置中，Pi0.5 取得了整体最优/领先的表现。综合来看，这些结果印证了 RoboTidy Benchmark 提供的多样化、背景丰富的轨迹能够弥补现有家庭场景数据的覆盖不足，从而在域移位条件下提升模型的泛化能力与鲁棒性。
\Cref{tab:VLA} presents results across four action tasks. Non-pretrained model (ACT~\cite{ACT}) performs poorly on unseen househols scenes and objects; in contrast, the pretrained models (RDT~\cite{RDT}, \(\pi_{0.5}~\cite{pi0.5}\)) maintain stronger robustness across different household scenes and objects, indicating that VLA pretraining provides effective priors for cross-domain generalization. Under our evaluation setup, \(\pi_{0.5}~\cite{pi0.5}\) achieves the best overall performance. Overall, these findings corroborate that the diverse, background-rich trajectories provided by the RoboTidy Benchmark complement the limited coverage of existing household datasets, thereby improving generalization and robustness.

\subsection{Navigation Task}

\noindent\textbf{Setting details.} For each household scene, we consider bidirectional navigation tasks between three rooms (\textit{living room}, \textit{kitchen} and \textit{bedroom}). The robot is equipped with an onboard, egocentric RGB camera with resolution $640 \times 480$. Each episode is automatically terminated if any of the following occurs: (1) the simulation time reaches 120 seconds; (2) the robot remains stationary beyond a predefined timeout or becomes unstable;  (3) the robot reaches the goal region. We fine-tune on 300 household scenes (0.9k trajectory--instruction pairs) and evaluate on the remaining scenes. Our experiments consider two backbones: Navid~\cite{NaVid} (navid-7b-full-224) and NaVILA~\cite{NaVILA} (navila-siglip-llama3-8b-v1.5-pretrain). We refer to the pretrained checkpoints as \textbf{Navid-base} and \textbf{NaVILA-base}, and to their RoboTidy-finetuned counterparts as \textbf{Navid-R} and \textbf{NaVILA-R} (``\textbf{R}'' denotes RoboTidy). We report navigation metrics: Success Rate (\textbf{SR}), Oracle Success Rate (\textbf{OSR}), and Success weighted by Path Length (\textbf{OSR}).

% 对于每一种家居场景，我们考虑在客厅-卧室，卧室—厨房，厨房-客厅这三个空间对之间的往返导航任务。每个回合中，当触发下列三种情况时，该回合自动终止：(1)仿真时间达到50秒；(2)机器人长时间停滞或因失衡无法继续前进；(3)机器人到达目标区域。我们从 RoboTidy 数据集中选取了 x对“轨迹—指令”样本，并在该子集上训练了两种模型：Navid-RoboTidy（以 Navid 的预训练模型 navid-7b-full-224 为初始化，记作 Navid-base）和 NaVILA-RoboTidy（以 NaVILA 的预训练模型 navila-siglip-llama3-8b-v1.5-pretrain 为初始化，记作 NaVILA-base）。

\noindent\textbf{Evaluation results.} As shown in \Cref{tab:vln}, we report results for several MLLMs and VLN models on our RoboTidy benchmark. Apart from NaVILA~\cite{NaVILA}, the recent SOTA VLN model, the SR values of the remaining models are all below 0.2. Taking NaVid-base as an example, the model achieves 0.22 SR on VLN-CE~\cite{VLNCE} R2R Val-Unseen, yet on our benchmark its SR decreases to 0.16. The results demonstrate that the proposed RoboTidy benchmark presents a more challenging setting for current VLN models. \Cref{tab:vlnce_results} reports the VLN-CE results of NaVid-R and NaVILA-R trained exclusively on our datasets. Relative to the baselines, the two models achieve marked improvements in robustness, highlighting the advantages brought by our diverse 3DGS household scenes.
%（后面开始说结论）：如表5所示, 我们展示了几种MLLMs和VLN models在our benchmark上的测试结果。除了近期SOTA的VLN model NaVILA,其余模型的成功率(SR)均在0.2以下。以NaVid-base为例，该模型在VLN-CE R2R Val-Unseen 上取得了SR 0.22，但在我们的benchmark上SR下降了约50%。这一结果说明了our RoboTidy Benchmark为当前的VLN-models提供了更有挑战性的任务场景。Table 6 展示了NaVid-R和NaVILA这两个仅在我们的数据上训练的模型在VLN-CE benchmark上的表现.与其他baselines比较之下，我们的两个模型在鲁棒性和效率方面均取得了显著提升，重点突出了我们多样化的3DGS家居场景所带来的优势。
%Relative to the baseline, X achieves marked improvements in both robustness and efficiency, highlighting the advantages brought by our RoboTidy Benchmark.

\begin{figure*}[t]
  \centering
  \includegraphics[width=0.99\textwidth]{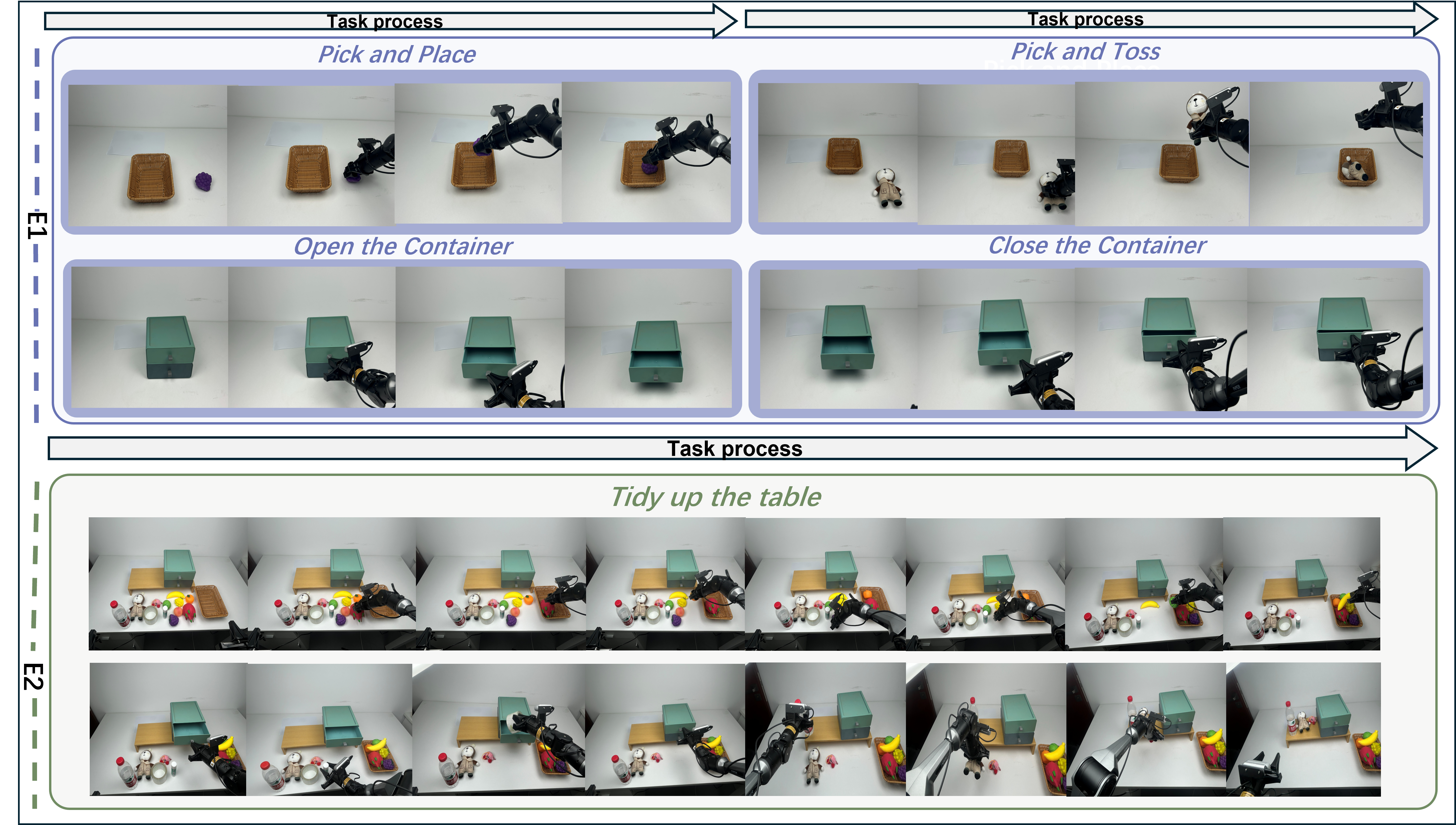}
  \caption{\textbf{Visualization of real-world tasks.}
\textbf{E1}: four manipulation action tasks \textit{(Pick and Place, Pick and Toss, Open the Container and Close the Container)}.
\textbf{E2}: household tidying task where the policy follows an \textit{"Action (Object, Container)" list} to sort objects. Task progresses from left to right. }
  \label{fig:E1_and_E2}
\end{figure*}

\subsection{Sim2Real Transfer Experiment}
We investigate a key question: \textit{what extent can robot trained in \textbf{RoboTidy} generalize to the real world?}
To evaluate Sim2Real transfer, we conduct two real-world manipulation experiments: \textbf{(E1)} under a fixed setting with one object and one container, evaluating four manipulation actions: \textit{Pick and Place}, \textit{Pick and Toss}, \textit{Open the Container} and \textit{Close the Container}; \textbf{(E2)} workbench multi-object sorting, performing a household tidying task  with 12 objects and 3 containers.

\begin{figure}[H]  
    \centering
    \includegraphics[width=0.99\linewidth]{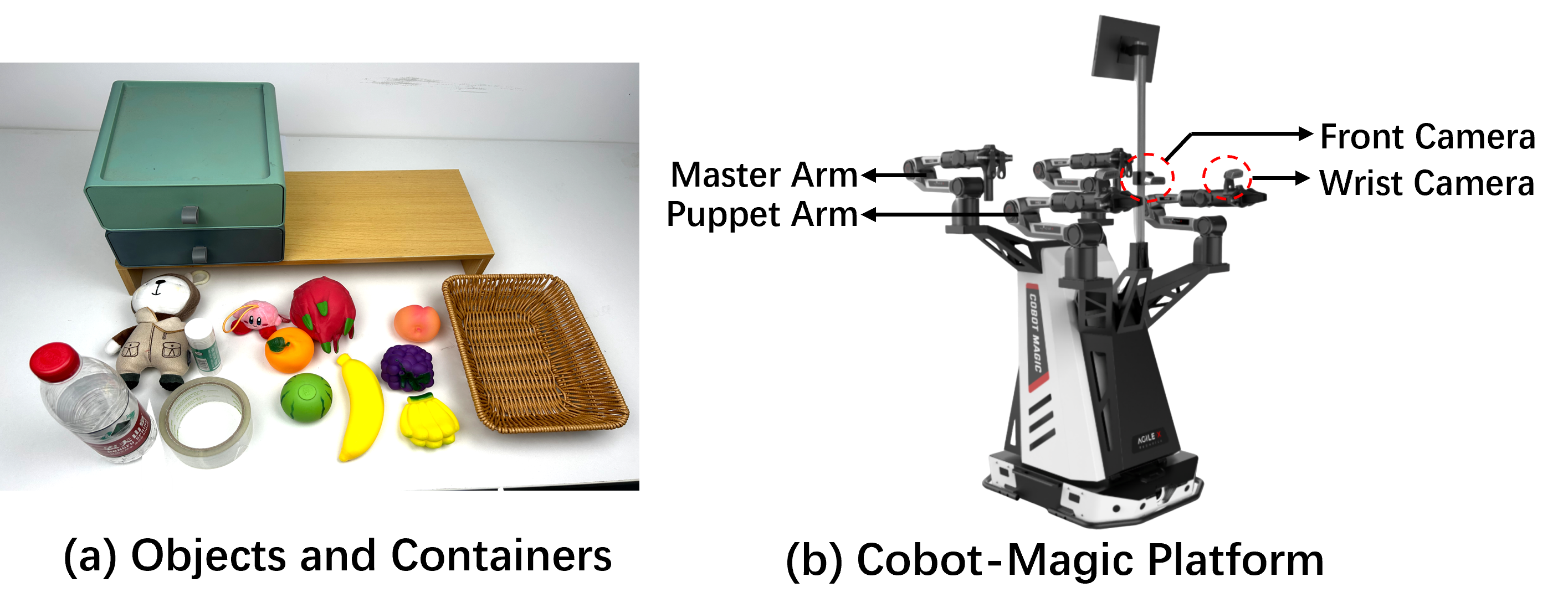} 
    \caption{\textbf{Real-world Experimental setup.} (a) Workspace with objects and containers. (b) Cobot-Magic dual-arm mobile manipulation platform.}
    \label{fig:platform_setup} 
\end{figure}

\noindent\textbf{Setting details}. 
%我们开展两项真实世界操控：(E1)针对manip[ualtion action，在单物体+单容器的固定场景中，仅检验 Pick and Place 和 pick and toss两种动作 (E2)为操作台多物体收纳，在包含 7 个物体与 2 个容器 的场景中  进行收纳任务我们会比较三种设定：(1)100real (2) 100real + 100Robotidy (3)100robotidy 两个试验均采用 RDT~\cite{RDT} 作为操控的 VLA 策略，并以 Qwen2-VL-72B~\cite{qwen2.5VL} 作为物品收纳的VLM backbone。。针对实验1我们选取任务成功率report ，针对试验2我们report收纳成功率。
As shown in \Cref{fig:platform_setup}, we set up a real-world testbed using a Cobot-Magic dual-arm mobile platform equipped with four Piper manipulators (two masters and two puppets). The system captures RGB images at 480×640 resolution at 30 Hz. We compare three training settings: (1) 50 real-world demonstrations. (2) a combined set of 50 real-world and 100 RoboTidy synthetic demonstrations \textbf{(FS)}. (3) 100 RoboTidy synthetic demonstrations only \textbf{(ZS)}. Both experiments adopt RDT~\cite{RDT} as the VLA policy for manipulation and use Qwen2.5-VL~\cite{qwen2.5VL} as the VLM backbone for object sorting. For \textbf{(E1)} we evaluate each manipultion action 10 times per setting and report \textbf{SR}, and for \textbf{(E2)} we report \textbf{SR} and \textbf{VSSR}.

\begin{figure}[H]  
    \centering
    \includegraphics[width=0.99\linewidth]{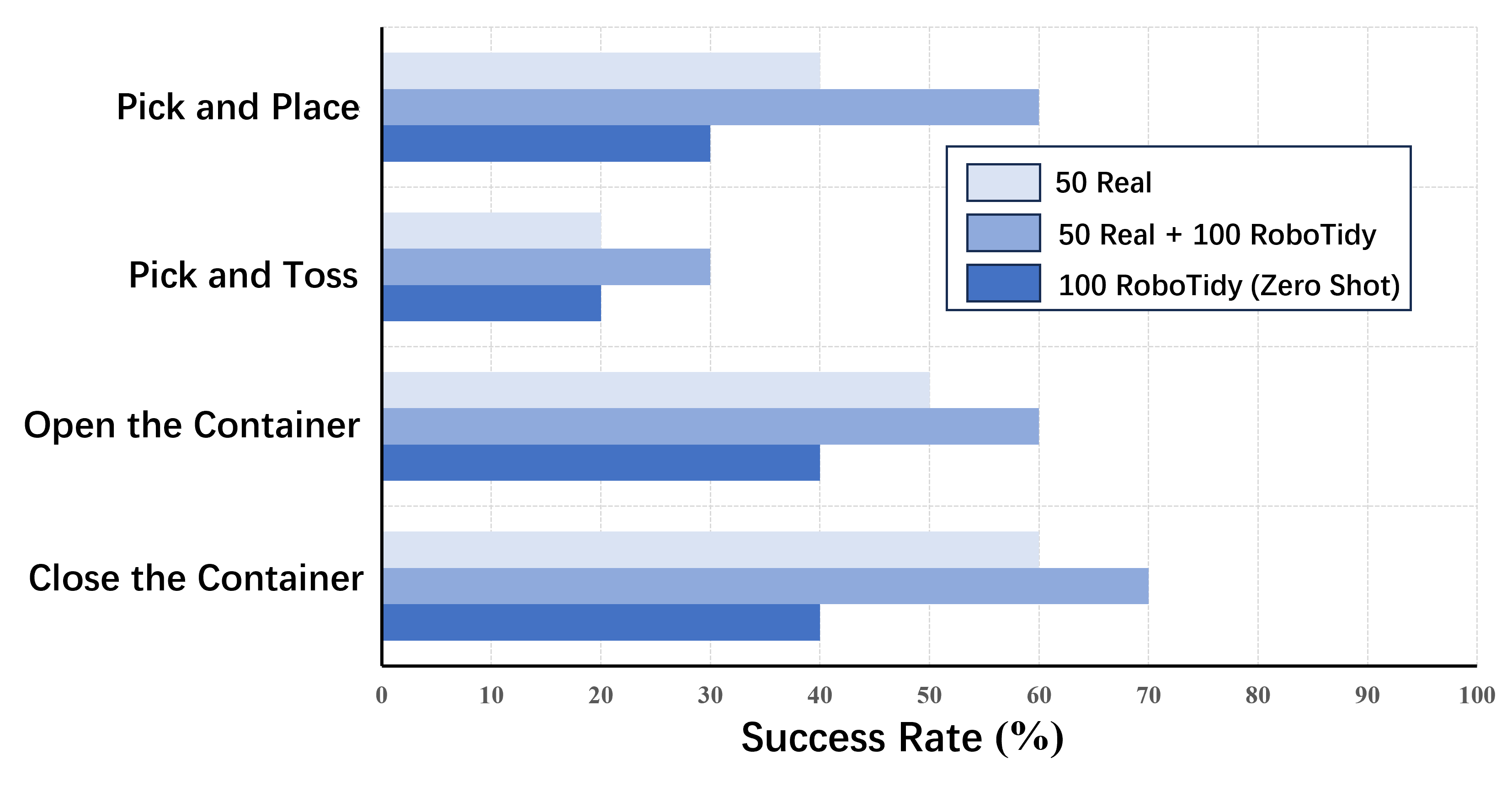} 
    \caption{\textbf{Real-world experimental results (E1).} We report success rates for four manipulation action tasks under three different settings.}
    \label{fig5:E1_four_manipulation_actions_tasks} 
\end{figure}

\noindent\textbf{Evaluation results}.
%实验结果表明：引入 RoboTidy 的合成示范轨迹可显著提升真实世界操控策略的成功率。在包含 100 条专家轨迹 的设置下，与仅使用 50 条真实示范 相比，平均成功率得到明显改善，其中 Pick and Place 提升约 20%，Pick and Toss open the container 和close the container提升 10%。包括在真实世界的场景中，收纳成功率也都有明显提升。在仅使用 RoboTidy 合成数据训练的 zero-shot 情况下，策略仍保持可观性能，尤其在 Pick and Toss 上与仅用 50 条真实数据的结果相当。在真实世界中，和真实数据仍有一定差距，但是差不太多(一个是5/12，另外是4/12，然后你看看这个话怎么说好一点呢)上述结果表明，这些结果表明，RoboTidy 提供的3DGS 场景的具有较高的视觉和物理仿真度从而促进可靠的 sim-to-real 迁移
As shown in \Cref{fig5:E1_four_manipulation_actions_tasks} and \Cref{tab:E2}, incorporating RoboTidy synthetic demonstration trajectories into training substantially improves the real-world manipulation success rate. With 100 RoboTidy demonstration trajectories, the \textbf{FS} setting outperforms training on 50 real-world demonstrations alone, yielding gains of approximately 20\% on \textit{Pick and Place} and 10\% on \textit{Pick and Toss}, \textit{Open the Container} and \textit{Close the Container}. Real-world stowing success also increases across the board. In the \textbf{ZS} setting trained solely on RoboTidy demonstrations, the policy still maintains competitive performance, matching the 50 real-world demonstrations on \textit{Pick and Toss}. In real-world evaluation, the \textbf{ZS} setting reaches 4/12 overall successes, versus 5/12 for the \textbf{FS} setting, and matches the 50 real-world demonstrations, indicating a small performance gap. Overall, these findings indicate that RoboTidy’s physics-consistent simulation and diverse object and container configurations effectively narrow the sim2real gap. These results indicate that our 3DGS household scenes provide photorealistic visual fidelity and physical consistency, thereby enabling reliable sim2real transfer.

\begin{table}[H]
\centering
\small
\caption{\textbf{Real-world experimental results (E2).} We report \textbf{SR} and \textbf{VSSR} for household tidying task under three different settings.}
\label{tab:E2}
\begin{tabular}{lccc}
\toprule
 Metric & 50 Real & 50 Real + 100 RoboTidy & 100 RoboTidy\\
\midrule
 SR   & 5/12 & 8/12 & 4/12 \\
 VSSR & 5/12 & 7/12 & 4/12 \\
\bottomrule
\end{tabular}
\end{table}

\subsection{Ablation Studies}
\noindent\textbf{Setting details.}
% 我们以“指令是否要求分类收纳（Sorting vs. No-Sorting）”作为唯一自变量进行消融。
% 对于 VLA（以 RDT 为策略）除语言指令外其它全部保持一致：感知输入、候选容器集合、动作原子、控制频率与重试策略。
% 仿真环境下评估 ZS 和 FS（FS 提供 2--3 个物体→容器示例）；真机端不提供任何示例（严格 ZS）。
% 指标为 \textbf{SR} 与 \textbf{VSSR}。
We conduct an ablation that isolates instruction semantics as the independent variable: one variant requires object sorting and the other does not. We use RDT~\cite{RDT} as the VLA policy and keep all other components fixed. In simulation, we perform tidying tasks in three rooms in each of ten unseen household scenes and evaluate under the \textbf{ZS} and \textbf{FS} settings. The real-world experiments follow the same setup as \textbf{E2} and are evaluated under the \textbf{ZS} and \textbf{FS} settings.

\begin{table*}[t]
\centering
\small
\setlength{\tabcolsep}{6pt}
\caption{Ablation with and without object sorting under \textbf{ZS} and \textbf{FS} settings. In simulation, we report \textbf{SR(\%)} and \textbf{VSSR(\%)} as the mean and standard deviation over 3 random seeds; in the real-world, results are counts over 12 objects and 3 containers.}
\label{tab:vla_ablation_fullwidth}
\begin{tabular}{lcccccccc}
\toprule
\multirow{2}{*}{Method}
& \multicolumn{2}{c}{Simulation \textbf{(ZS)}}
& \multicolumn{2}{c}{Simulation \textbf{(FS)}}
& \multicolumn{2}{c}{Real-world \textbf{(ZS)}}
& \multicolumn{2}{c}{Real-world \textbf{(FS)}} \\
& SR & VSSR & SR & VSSR & SR & VSSR & SR & VSSR \\
\midrule
RDT~\cite{RDT}                & 
68.4{\color{gray!80}\footnotesize$\,\pm\,1.4$}&
59.3{\color{gray!80}\footnotesize$\,\pm\,0.6$}&
70.1{\color{gray!80}\footnotesize$\,\pm\,0.9$}& 
63.2{\color{gray!80}\footnotesize$\,\pm\,1.1$}& 
4/12  & 7/12  & 5/12 & 7/12 \\
RDT w/ object sorting &
76.3{\color{gray!80}\footnotesize$\,\pm\,0.3$}&
70.8{\color{gray!80}\footnotesize$\,\pm\,1.4$}&
81.3{\color{gray!80}\footnotesize$\,\pm\,2.1$}& 
74.5{\color{gray!80}\footnotesize$\,\pm\,1.8$}&      
5/12 & 5/12 & 7/12 & 8/12 \\
\bottomrule
\end{tabular}
\label{tab:ablation_studies}
\end{table*}

\noindent\textbf{Evaluation results.}
%% 无论在仿真还是真机环境，去掉分类收纳约束都会使性能显著下降。
% 在仿真中，\textbf{SR} 平均下降约 10 个百分点，并且由于误放，\textbf{VSSR} 降幅更明显。
% 在真机端，这一退化更为突出，原因在于指令不明确。
% 结果表明：在指令中显式加入“按语义进行分类收纳”的要求，能显著提升容器选择的正确性并稳定后续操控，体现了 object sorting 对家庭整理任务的价值。
As shown in \Cref{tab:ablation_studies}, altering only the instruction semantics by requiring object sorting yields consistent gains. In simulation \textbf{ZS} setting, SR (+7.9\%) and VSSR (+11.5\%); in simulation \textbf{FS} setting, SR (+11.2\%) and VSSR (+11.3\%). In the real world, both \textbf{ZS} and \textbf{FS} settings also show improvements, indicating that clear sorting rules shrink the search space under perception noise and reduce failures. Overall, removing the object sorting rules degrades performance; making object sorting rules explicit at the instruction level is key to accurate container selection and higher end to end success, with the same trend under both \textbf{ZS} and \textbf{FS} settings.

\section{Conclusion}

%本文提出 RoboTidy，一个面向家居收纳的统一基准，覆盖 VLA 与 VLN 两条训练与评测链路。我们将“收纳”形式化为 Action(Object, Container) 的结构化列表，并提供四类操控原子动作、具有物理可执行性的 3DGS 家居场景与多模态感知，以支持闭环执行与大规模数据采集。RoboTidy 基于 500 个室内场景，包含 6k 条操控示范与 1.5k 条跨房间导航轨迹；在此之上我们系统评测代表性方法，并开展真实世界部署以验证从仿真到现实的可迁移性。实验表明：强大的 VLM 在零样本与小样本条件下对对象归类与容器推断具有良好泛化；RoboTidy 在导航任务上更具挑战；预训练 VLA 在未见场景与对象上表现出更高稳健性；在本基准上微调的 VLN 模型在鲁棒性与效率方面均获得稳定提升。
We introduced RoboTidy, a language-guided benchmark for household tidying that evaluates embodied agents in both VLA and VLN. The benchmark combines photorealistic 3DGS household scenes with physics collisions, manipulation demonstrations and cross-room navigation trajectories, and an \textit{"Action (Object, Container)" list} that captures user-interpretable placement preferences. It also provides language instructions and standardized evaluation metrics, enabling fair and consistent comparison. Building on this design, we conducted representative baselines for VLA and VLN and validated sim2real transfer in the real world. RoboTidy supports sim2real transfer and emphasizes reproducibility and extensibility. We expect RoboTidy to serve as a foundation for future work on richer long horizon actions and  interactive manipulation, and broader sim2real studies.
%We present RoboTidy, a language-guided benchmark for household tidying that support VLA and VLN training and evaluation. Household tidying is formalized as a structured \textit{“Action (Object, Container)” list}, accompanied by four manipulation actions, physically executable 3DGS household scenes, and multimodal sensing to enable closed-loop execution and scalable data collection. Built on 500 household scenes, RoboTidy includes 6.4k manipulation demonstrations and 1.5k cross-room navigation trajectories. On this basis, we conduct systematic evaluations of representative methods and deploy in the real world to validate sim-to-real transfer. Experiments show that strong VLM generalize well for object categorization and container inference in ~\textbf{ZS} and ~\textbf{FS} settings; RoboTidy is notably more challenging for navigation; pretrained VLA policies are more robust in unseen scenes and objects; and VLN models fine-tuned on RoboTidy achieve consistent gains in robustness and efficiency.
%首先就是说我们专注逼真的仿真场景，但是我们的仍然与真实世界具有一定的差距，首先就是我们后续会添加更多的物体以及更多的场景来缩小这部分差距，其次由于真实环境下灯光等变换复杂，我们在真机试验中没有对背景进行过于复杂的变化。第三我们后续会在语言指令上进一步丰富，以及未来我们也会继续扩大示范数据，这能够极大的推动具身智能领域的发展，差不多这个意思。
%尽管我们专注于构建照片级的仿真环境，当前系统与真实世界仍存在差距；后续将进一步扩充物体与场景的广度与多样性。在真实实验中，由于光照与背景变化的复杂性，部分复杂家庭布局尚未覆盖。未来我们将进一步丰富操控动作集合，并持续扩大示范数据规模，以提升模型在开放环境中的鲁棒性与泛化能力。同时，我们将系统性纳入跨具身体与多相机配置下的评测协议，覆盖更长时序的跨房间任务与更强遮挡条件，增强对用户收纳偏好的表达与在线建模能力。我们还将补充动态人类干扰与移动障碍的场景，改进闭环失败恢复与安全约束策略，并在材质与光谱一致性、传感器噪声与时延建模方面缩小仿真—真实差距。数据、资产与基线实现将继续开放并按即插即用方式组织，以支持可复现实验与社区对比研究。

\noindent\textbf{Limitations and Future Work.} Although we focus on building photorealistic simulated household environments, a gap to the real world remains. We will expand the breadth of objects and scenes. In our real world experiments, the complexity of lighting and background variation meant that some challenging household scenes were not covered. Going forward, we will enrich the set of manipulation actions and continue to scale up demonstration data to improve robustness and generalization in more challenging environments. We will also incorporate scenarios with human interference and moving obstacles to evaluate policies under dynamic conditions. These additions will stress test closed loop recovery and safety and bring the benchmark closer to real deployments, thereby advancing embodied intelligence more effectively. In addition, we will apply stronger domain randomization for illumination, textures, and clutter, and we will report standardized diagnostics of failure modes to guide future research.
%Although we focus on building photorealistic simulation household environments, a gap to the real world remains; we will expand the breadth of objects and scenes going forward. In real-world experiments, due to the complexity of lighting and background variations, some challenging household scenes were not covered. In the future, we will further enrich the manipulation actions and continue to scale up demonstration data to improve robustness and generalization in  more challenging environments, We will also incorporate scenarios with human interference and moving obstacles to evaluate policies under dynamic conditions. These additions will stress test closed-loop recovery and safety, bringing the benchmark closer to real deployments. thereby more effectively advancing embodied intelligence.
{
    \small
    \bibliographystyle{ieeenat_fullname}
    \bibliography{main}

\begin{thebibliography}{47}
\providecommand{\natexlab}[1]{#1}
\providecommand{\url}[1]{\texttt{#1}}
\expandafter\ifx\csname urlstyle\endcsname\relax
  \providecommand{\doi}[1]{doi: #1}\else
  \providecommand{\doi}{doi: \begingroup \urlstyle{rm}\Url}\fi

\bibitem[Bai et~al.(2025)Bai, Chen, Liu, Wang, Ge, Song, Dang, Wang, Wang, Tang, et~al.]{qwen2.5VL}
Shuai Bai, Keqin Chen, Xuejing Liu, Jialin Wang, Wenbin Ge, Sibo Song, Kai Dang, Peng Wang, Shijie Wang, Jun Tang, et~al.
\newblock Qwen2. 5-vl technical report.
\newblock \emph{arXiv preprint arXiv:2502.13923}, 2025.

\bibitem[Brohan et~al.(2023)Brohan, Brown, Carbajal, Chebotar, Dabis, Finn, , et~al.]{RDT}
Anthony Brohan, Noah Brown, Justice Carbajal, Yevgen Chebotar, Joseph Dabis, Chelsea Finn, , et~al.
\newblock Rt-1: Robotics transformer for real-world control at scale.
\newblock \emph{Proceedings of Robotics: Science and Systems}, 2023.

\bibitem[Chang et~al.(2017)Chang, Dai, Funkhouser, Halber, Niebner, Savva, Song, Zeng, and Zhang]{matterport3d}
Angel Chang, Angela Dai, Thomas Funkhouser, Maciej Halber, Matthias Niebner, Manolis Savva, Shuran Song, Andy Zeng, and Yinda Zhang.
\newblock Matterport3d: Learning from rgb-d data in indoor environments.
\newblock In \emph{2017 International Conference on 3D Vision (3DV)}, pages 667--676. IEEE Computer Society, 2017.

\bibitem[Chen et~al.(2023)Chen, Tamboli, Lan, and Aggarwal]{DBLP:conf/icml/ChenTLA23}
Jiayu Chen, Dipesh Tamboli, Tian Lan, and Vaneet Aggarwal.
\newblock Multi-task hierarchical adversarial inverse reinforcement learning.
\newblock In \emph{International Conference on Machine Learning}, pages 4895--4920. {PMLR}, 2023.

\bibitem[Chen et~al.(2025)Chen, Chen, Chen, Cai, Liu, Li, Liang, Lin, Ge, Gu, et~al.]{robotwin}
Tianxing Chen, Zanxin Chen, Baijun Chen, Zijian Cai, Yibin Liu, Zixuan Li, Qiwei Liang, Xianliang Lin, Yiheng Ge, Zhenyu Gu, et~al.
\newblock Robotwin 2.0: A scalable data generator and benchmark with strong domain randomization for robust bimanual robotic manipulation.
\newblock \emph{arXiv preprint arXiv:2506.18088}, 2025.

\bibitem[Chen et~al.(2024)Chen, Wang, Cao, Liu, Gao, Cui, Zhu, Ye, Tian, Liu, et~al.]{InternVL}
Zhe Chen, Weiyun Wang, Yue Cao, Yangzhou Liu, Zhangwei Gao, Erfei Cui, Jinguo Zhu, Shenglong Ye, Hao Tian, Zhaoyang Liu, et~al.
\newblock Expanding performance boundaries of open-source multimodal models with model, data, and test-time scaling.
\newblock \emph{arXiv preprint arXiv:2412.05271}, 2024.

\bibitem[Cheng et~al.(2024)Cheng, Ji, Yang, Gongye, Zou, Kautz, B{\i}y{\i}k, Yin, Liu, and Wang]{NaVILA}
An-Chieh Cheng, Yandong Ji, Zhaojing Yang, Zaitian Gongye, Xueyan Zou, Jan Kautz, Erdem B{\i}y{\i}k, Hongxu Yin, Sifei Liu, and Xiaolong Wang.
\newblock Navila: Legged robot vision-language-action model for navigation.
\newblock \emph{arXiv preprint arXiv:2412.04453}, 2024.

\bibitem[Choi et~al.(2024)Choi, Yoon, Ong, Kim, and Jang]{lota-bench}
Jae-Woo Choi, Youngwoo Yoon, Hyobin Ong, Jaehong Kim, and Minsu Jang.
\newblock Lota-bench: Benchmarking language-oriented task planners for embodied agents.
\newblock \emph{arXiv preprint arXiv:2402.08178}, 2024.

\bibitem[Dubey et~al.(2024)Dubey, Jauhri, Pandey, Kadian, Al-Dahle, Letman, Mathur, Schelten, Yang, Fan, et~al.]{Llama3}
Abhimanyu Dubey, Abhinav Jauhri, Abhinav Pandey, Abhishek Kadian, Ahmad Al-Dahle, Aiesha Letman, Akhil Mathur, Alan Schelten, Amy Yang, Angela Fan, et~al.
\newblock The llama 3 herd of models.
\newblock \emph{arXiv e-prints}, pages arXiv--2407, 2024.

\bibitem[Ehsani et~al.(2021)Ehsani, Han, Herrasti, VanderBilt, Weihs, Kolve, Kembhavi, and Mottaghi]{manipulathor}
Kiana Ehsani, Winson Han, Alvaro Herrasti, Eli VanderBilt, Luca Weihs, Eric Kolve, Aniruddha Kembhavi, and Roozbeh Mottaghi.
\newblock Manipulathor: A framework for visual object manipulation.
\newblock In \emph{Proceedings of the IEEE/CVF conference on computer vision and pattern recognition}, pages 4497--4506, 2021.

\bibitem[Gong et~al.(2023)Gong, Huang, Zhao, Geng, Gao, Wu, Ai, Zhou, Terzopoulos, Zhu, et~al.]{arnold}
Ran Gong, Jiangyong Huang, Yizhou Zhao, Haoran Geng, Xiaofeng Gao, Qingyang Wu, Wensi Ai, Ziheng Zhou, Demetri Terzopoulos, Song-Chun Zhu, et~al.
\newblock Arnold: A benchmark for language-grounded task learning with continuous states in realistic 3d scenes.
\newblock In \emph{Proceedings of the IEEE/CVF International Conference on Computer Vision}, pages 20483--20495, 2023.

\bibitem[Gu et~al.(2023)Gu, Xiang, Li, Ling, Liu, Mu, Tang, Tao, Wei, Yao, et~al.]{maniskill2}
Jiayuan Gu, Fanbo Xiang, Xuanlin Li, Zhan Ling, Xiqiang Liu, Tongzhou Mu, Yihe Tang, Stone Tao, Xinyue Wei, Yunchao Yao, et~al.
\newblock Maniskill2: A unified benchmark for generalizable manipulation skills.
\newblock \emph{arXiv preprint arXiv:2302.04659}, 2023.

\bibitem[Hart et~al.(1968)Hart, Nilsson, and Raphael]{A*}
Peter~E Hart, Nils~J Nilsson, and Bertram Raphael.
\newblock A formal basis for the heuristic determination of minimum cost paths.
\newblock \emph{IEEE transactions on Systems Science and Cybernetics}, 4\penalty0 (2):\penalty0 100--107, 1968.

\bibitem[Intelligence et~al.(2025)Intelligence, Black, Brown, Darpinian, Dhabalia, Driess, Esmail, Equi, Finn, Fusai, et~al.]{pi0.5}
Physical Intelligence, Kevin Black, Noah Brown, James Darpinian, Karan Dhabalia, Danny Driess, Adnan Esmail, Michael Equi, Chelsea Finn, Niccolo Fusai, et~al.
\newblock $\pi_{0.5}$: A {Vision–Language–Action} model with {Open-World} generalization.
\newblock \emph{arXiv preprint arXiv:2504.16054}, 2025.

\bibitem[James et~al.(2020)James, Ma, Arrojo, and Davison]{rlbench}
Stephen James, Zicong Ma, David~Rovick Arrojo, and Andrew~J Davison.
\newblock Rlbench: The robot learning benchmark \& learning environment.
\newblock \emph{IEEE Robotics and Automation Letters}, 5\penalty0 (2):\penalty0 3019--3026, 2020.

\bibitem[Kant et~al.(2022)Kant, Ramachandran, Yenamandra, Gilitschenski, Batra, Szot, and Agrawal]{housekeep}
Yash Kant, Arun Ramachandran, Sriram Yenamandra, Igor Gilitschenski, Dhruv Batra, Andrew Szot, and Harsh Agrawal.
\newblock Housekeep: Tidying virtual households using commonsense reasoning.
\newblock In \emph{European Conference on Computer Vision}, pages 355--373. Springer, 2022.

\bibitem[Kerbl et~al.(2023)Kerbl, Kopanas, Leimk{\"u}hler, and Drettakis]{3dgs}
Bernhard Kerbl, Georgios Kopanas, Thomas Leimk{\"u}hler, and George Drettakis.
\newblock 3d gaussian splatting for real-time radiance field rendering.
\newblock \emph{ACM Trans. Graph.}, 42\penalty0 (4):\penalty0 139--1, 2023.

\bibitem[Krantz et~al.(2020{\natexlab{a}})Krantz, Wijmans, Majumdar, Batra, and Lee]{VLNCE}
Jacob Krantz, Erik Wijmans, Arjun Majumdar, Dhruv Batra, and Stefan Lee.
\newblock Beyond the nav-graph: Vision-and-language navigation in continuous environments.
\newblock In \emph{European Conference on Computer Vision}, pages 104--120. Springer, 2020{\natexlab{a}}.

\bibitem[Krantz et~al.(2020{\natexlab{b}})Krantz, Wijmans, Majumdar, Batra, and Lee]{seq2seq}
Jacob Krantz, Erik Wijmans, Arjun Majumdar, Dhruv Batra, and Stefan Lee.
\newblock Beyond the nav-graph: Vision-and-language navigation in continuous environments.
\newblock In \emph{European Conference on Computer Vision}, pages 104--120. Springer, 2020{\natexlab{b}}.

\bibitem[Li et~al.(2023)Li, Zhang, Wong, Gokmen, Srivastava, Mart{\'\i}n-Mart{\'\i}n, Wang, Levine, Lingelbach, Sun, et~al.]{Behavior-1k}
Chengshu Li, Ruohan Zhang, Josiah Wong, Cem Gokmen, Sanjana Srivastava, Roberto Mart{\'\i}n-Mart{\'\i}n, Chen Wang, Gabrael Levine, Michael Lingelbach, Jiankai Sun, et~al.
\newblock Behavior-1k: A benchmark for embodied ai with 1,000 everyday activities and realistic simulation.
\newblock In \emph{Conference on Robot Learning}, pages 80--93. PMLR, 2023.

\bibitem[Liu et~al.(2023)Liu, Zhu, Gao, Feng, Liu, Zhu, and Stone]{libero}
Bo Liu, Yifeng Zhu, Chongkai Gao, Yihao Feng, Qiang Liu, Yuke Zhu, and Peter Stone.
\newblock Libero: Benchmarking knowledge transfer for lifelong robot learning.
\newblock \emph{arXiv preprint arXiv:2306.03310}, 2023.

\bibitem[Liu et~al.(2019)Liu, Ott, Goyal, Du, Joshi, Chen, Levy, Lewis, Zettlemoyer, and Stoyanov]{Roberta}
Yinhan Liu, Myle Ott, Naman Goyal, Jingfei Du, Mandar Joshi, Danqi Chen, Omer Levy, Mike Lewis, Luke Zettlemoyer, and Veselin Stoyanov.
\newblock Roberta: A robustly optimized bert pretraining approach.
\newblock \emph{arXiv preprint arXiv:1907.11692}, 2019.

\bibitem[Long et~al.(2024)Long, Cai, Wang, Zhan, and Dong]{instructnav}
Yuxing Long, Wenzhe Cai, Hongcheng Wang, Guanqi Zhan, and Hao Dong.
\newblock Instructnav: Zero-shot system for generic instruction navigation in unexplored environment.
\newblock \emph{arXiv preprint arXiv:2406.04882}, 2024.

\bibitem[Mees et~al.(2022)Mees, Hermann, Rosete-Beas, and Burgard]{calvin}
Oier Mees, Lukas Hermann, Erick Rosete-Beas, and Wolfram Burgard.
\newblock Calvin: A benchmark for language-conditioned policy learning for long-horizon robot manipulation tasks.
\newblock \emph{IEEE Robotics and Automation Letters (RA-L)}, 7\penalty0 (3):\penalty0 7327--7334, 2022.

\bibitem[Miao et~al.(2025)Miao, Wei, Ge, Sun, Gao, Zhu, Wang, Tang, Xiao, Tang, Li, et~al.]{miao}
Bingchen Miao, Rong Wei, Zhiqi Ge, Xiaoquan Sun, Shiqi Gao, Jingzhe Zhu, Renhan Wang, Siliang Tang, Jun Xiao, Rui Tang, Juncheng Li, et~al.
\newblock Towards physically executable 3d gaussian for embodied navigation.
\newblock \emph{arXiv preprint arXiv:2510.21307}, 2025.

\bibitem[{NVIDIA}(2025)]{NVIDIA_Isaac_Sim}
{NVIDIA}.
\newblock {Isaac Sim}.
\newblock \url{https://github.com/isaac-sim/IsaacSim}, 2025.

\bibitem[Padmakumar et~al.(2022)Padmakumar, Thomason, Shrivastava, Lange, Narayan-Chen, Gella, Piramuthu, Tur, and Hakkani-Tur]{teach}
Aishwarya Padmakumar, Jesse Thomason, Ayush Shrivastava, Patrick Lange, Anjali Narayan-Chen, Spandana Gella, Robinson Piramuthu, Gokhan Tur, and Dilek Hakkani-Tur.
\newblock Teach: Task-driven embodied agents that chat.
\newblock In \emph{Proceedings of the AAAI Conference on Artificial Intelligence}, pages 2017--2025, 2022.

\bibitem[Puig et~al.(2018)Puig, Ra, Boben, Li, Wang, Fidler, and Torralba]{virtualhome}
Xavier Puig, Kevin Ra, Marko Boben, Jiaman Li, Tingwu Wang, Sanja Fidler, and Antonio Torralba.
\newblock Virtualhome: Simulating household activities via programs.
\newblock In \emph{2018 IEEE/CVF Conference on Computer Vision and Pattern Recognition}. IEEE, 2018.

\bibitem[Pumacay et~al.()Pumacay, Singh, Duan, Krishna, Thomason, and Fox]{colosseum}
Wilbert Pumacay, Ishika Singh, Jiafei Duan, Ranjay Krishna, Jesse Thomason, and Dieter Fox.
\newblock The colosseum: A benchmark for evaluating generalization for robotic manipulation.
\newblock In \emph{RSS 2024 Workshop: Data Generation for Robotics}.

\bibitem[Radford et~al.(2021)Radford, Kim, Hallacy, Ramesh, Goh, Agarwal, Sastry, Askell, Mishkin, Clark, et~al.]{CLIP}
Alec Radford, Jong~Wook Kim, Chris Hallacy, Aditya Ramesh, Gabriel Goh, Sandhini Agarwal, Girish Sastry, Amanda Askell, Pamela Mishkin, Jack Clark, et~al.
\newblock Learning transferable visual models from natural language supervision.
\newblock In \emph{International conference on machine learning}, pages 8748--8763. PmLR, 2021.

\bibitem[Ramakrishnan et~al.()Ramakrishnan, Gokaslan, Wijmans, Maksymets, Clegg, Turner, Undersander, Galuba, Westbury, Chang, et~al.]{HM3D}
Santhosh~Kumar Ramakrishnan, Aaron Gokaslan, Erik Wijmans, Oleksandr Maksymets, Alexander Clegg, John~M Turner, Eric Undersander, Wojciech Galuba, Andrew Westbury, Angel~X Chang, et~al.
\newblock Habitat-matterport 3d dataset (hm3d): 1000 large-scale 3d environments for embodied ai.
\newblock In \emph{Thirty-fifth Conference on Neural Information Processing Systems Datasets and Benchmarks Track (Round 2)}.

\bibitem[Rasch et~al.(2019)Rasch, Sprute, P{\"o}rtner, Battermann, and K{\"o}nig]{tidy}
Robin Rasch, Dennis Sprute, Aljoscha P{\"o}rtner, Sven Battermann, and Matthias K{\"o}nig.
\newblock Tidy up my room: Multi-agent cooperation for service tasks in smart environments.
\newblock \emph{Journal of Ambient Intelligence and Smart Environments}, 11\penalty0 (3):\penalty0 261--275, 2019.

\bibitem[Reimers and Gurevych(2019)]{sentence}
Nils Reimers and Iryna Gurevych.
\newblock Sentence-bert: Sentence embeddings using siamese bert-networks.
\newblock In \emph{Proceedings of the 2019 Conference on Empirical Methods in Natural Language Processing and the 9th International Joint Conference on Natural Language Processing (EMNLP-IJCNLP)}, page 3982. Association for Computational Linguistics, 2019.

\bibitem[Schulman et~al.(2017)Schulman, Wolski, Dhariwal, Radford, and Klimov]{PPO}
John Schulman, Filip Wolski, Prafulla Dhariwal, Alec Radford, and Oleg Klimov.
\newblock Proximal policy optimization algorithms.
\newblock \emph{arXiv preprint arXiv:1707.06347}, 2017.

\bibitem[Shridhar et~al.(2020)Shridhar, Thomason, Gordon, Bisk, Han, Mottaghi, Zettlemoyer, and Fox]{alfred}
Mohit Shridhar, Jesse Thomason, Daniel Gordon, Yonatan Bisk, Winson Han, Roozbeh Mottaghi, Luke Zettlemoyer, and Dieter Fox.
\newblock Alfred: A benchmark for interpreting grounded instructions for everyday tasks.
\newblock In \emph{Proceedings of the IEEE/CVF conference on computer vision and pattern recognition}, pages 10740--10749, 2020.

\bibitem[SpatialVerse Research~Team(2025)]{InteriorGS}
Manycore Tech~Inc. SpatialVerse Research~Team.
\newblock Interiorgs: A 3d gaussian splatting dataset of semantically labeled indoor scenes.
\newblock \url{https://huggingface.co/datasets/spatialverse/InteriorGS}, 2025.

\bibitem[Szot et~al.(2021)Szot, Clegg, Undersander, Wijmans, Zhao, Turner, Maestre, Mukadam, Chaplot, Maksymets, et~al.]{habitat}
Andrew Szot, Alexander Clegg, Eric Undersander, Erik Wijmans, Yili Zhao, John Turner, Noah Maestre, Mustafa Mukadam, Devendra~Singh Chaplot, Oleksandr Maksymets, et~al.
\newblock Habitat 2.0: Training home assistants to rearrange their habitat.
\newblock \emph{Advances in neural information processing systems}, 34:\penalty0 251--266, 2021.

\bibitem[Wei et~al.(2022)Wei, Liu, Ling, and Su]{Wei2022CoACD}
Xinyue Wei, Minghua Liu, Zhan Ling, and Hao Su.
\newblock Approximate convex decomposition for 3d meshes with collision-aware concavity and tree search.
\newblock \emph{ACM Transactions on Graphics (TOG)}, 41\penalty0 (4):\penalty0 1--18, 2022.

\bibitem[Weihs et~al.(2021)Weihs, Deitke, Kembhavi, and Mottaghi]{visualroom}
Luca Weihs, Matt Deitke, Aniruddha Kembhavi, and Roozbeh Mottaghi.
\newblock Visual room rearrangement.
\newblock In \emph{Proceedings of the IEEE/CVF conference on computer vision and pattern recognition}, pages 5922--5931, 2021.

\bibitem[Wu et~al.(2023)Wu, Antonova, Kan, Lepert, Zeng, Song, Bohg, Rusinkiewicz, and Funkhouser]{tidybot}
Jimmy Wu, Rika Antonova, Adam Kan, Marion Lepert, Andy Zeng, Shuran Song, Jeannette Bohg, Szymon Rusinkiewicz, and Thomas Funkhouser.
\newblock Tidybot: Personalized robot assistance with large language models.
\newblock \emph{Autonomous Robots}, 47\penalty0 (8):\penalty0 1087--1102, 2023.

\bibitem[Wu et~al.(2025)Wu, Martinez~Esturo, Mirzaei, Moenne-Loccoz, and Gojcic]{wu20253dgut}
Qi Wu, Janick Martinez~Esturo, Ashkan Mirzaei, Nicolas Moenne-Loccoz, and Zan Gojcic.
\newblock 3dgut: Enabling distorted cameras and secondary rays in gaussian splatting.
\newblock \emph{Conference on Computer Vision and Pattern Recognition (CVPR)}, 2025.

\bibitem[Yan et~al.(2021{\natexlab{a}})Yan, Crombez, Buisson, Ruichck, Krajnik, and Sun]{kant}
Zhi Yan, Nathan Crombez, Jocelyn Buisson, Yassine Ruichck, Tomas Krajnik, and Li Sun.
\newblock A quantifiable stratification strategy for tidy-up in service robotics.
\newblock In \emph{2021 IEEE international conference on advanced robotics and its social impacts (ARSO)}, pages 182--187. IEEE, 2021{\natexlab{a}}.

\bibitem[Yan et~al.(2021{\natexlab{b}})Yan, Crombez, Buisson, Ruichck, Krajnik, and Sun]{taniguchi}
Zhi Yan, Nathan Crombez, Jocelyn Buisson, Yassine Ruichck, Tomas Krajnik, and Li Sun.
\newblock A quantifiable stratification strategy for tidy-up in service robotics.
\newblock In \emph{2021 IEEE international conference on advanced robotics and its social impacts (ARSO)}, pages 182--187. IEEE, 2021{\natexlab{b}}.

\bibitem[Yan et~al.(2021{\natexlab{c}})Yan, Crombez, Buisson, Ruichck, Krajnik, and Sun]{yan2021}
Zhi Yan, Nathan Crombez, Jocelyn Buisson, Yassine Ruichck, Tomas Krajnik, and Li Sun.
\newblock A quantifiable stratification strategy for tidy-up in service robotics.
\newblock In \emph{2021 IEEE international conference on advanced robotics and its social impacts (ARSO)}, pages 182--187. IEEE, 2021{\natexlab{c}}.

\bibitem[Zeng et~al.(2020)Zeng, Song, Lee, Rodriguez, and Funkhouser]{tossingbot}
Andy Zeng, Shuran Song, Johnny Lee, Alberto Rodriguez, and Thomas Funkhouser.
\newblock Tossingbot: Learning to throw arbitrary objects with residual physics.
\newblock \emph{IEEE Transactions on Robotics}, 36\penalty0 (4):\penalty0 1307--1319, 2020.

\bibitem[Zhang et~al.(2024)Zhang, Wang, Xu, Zhou, Hong, Fang, Wu, Zhang, and Wang]{NaVid}
Jiazhao Zhang, Kunyu Wang, Rongtao Xu, Gengze Zhou, Yicong Hong, Xiaomeng Fang, Qi Wu, Zhizheng Zhang, and He Wang.
\newblock Navid: Video-based vlm plans the next step for vision-and-language navigation.
\newblock \emph{arXiv preprint arXiv:2402.15852}, 2024.

\bibitem[Zhao et~al.(2023)Zhao, Kumar, Levine, and Finn]{ACT}
Tony~Z. Zhao, Vikash Kumar, Sergey Levine, and Chelsea Finn.
\newblock Learning fine-grained bimanual manipulation with low-cost hardware.
\newblock In \emph{Proceedings of Robotics: Science and Systems}, 2023.

\end{thebibliography}
}


\end{document}